\documentclass[runningheads]{llncs}

\usepackage{comment}
\usepackage{epsfig}
\usepackage{graphicx}
\usepackage{amsmath}
\usepackage{amssymb}
\usepackage{gensymb}
\usepackage{multirow}
\usepackage{enumitem}
\usepackage{subcaption}
\usepackage{xspace}
\usepackage{xstring}
\usepackage{placeins}
\usepackage{floatrow}
\usepackage{balance}
\usepackage{cite}  
\usepackage{url}
\usepackage{pgffor}
\usepackage[dvipsnames]{xcolor}
\graphicspath{{./images/}{./supp_images/}}

\usepackage[pagebackref=true,breaklinks=true,letterpaper=true,colorlinks,bookmarks=false]{hyperref}
\usepackage{cleveref}

\begin{document}
\pagestyle{headings}
\mainmatter
\def\ECCVSubNumber{1889}  

\newcommand{\etal}{\textit{et al}.}
\newcommand{\ie}{\textit{i}.\textit{e}. }
\newcommand{\eg}{\textit{e}.\textit{g}. }

\title{ContactPose: A Dataset of Grasps with Object Contact and Hand Pose}

\titlerunning{ContactPose: A Dataset of Grasps with Object Contact and Hand Pose}
%
\author{Samarth Brahmbhatt\inst{1}\orcidID{0000-0002-3732-8865} \and
Chengcheng Tang\inst{3} \and
Christopher D. Twigg\inst{3} \and
Charles C. Kemp\inst{1} \and
James Hays\inst{1,2}}
\authorrunning{S. Brahmbhatt et al.}
%
\institute{Georgia Tech, Atlanta GA, USA
\email{\{samarth.robo,hays\}@gatech.edu, charlie.kemp@bme.gatech.edu} \and
Argo AI \and
Facebook Reality Labs \email{\{chengcheng.tang,cdtwigg\}@fb.com}}
\maketitle

\begin{abstract}
Grasping is natural for humans. However, it involves complex hand configurations and soft tissue deformation that can result in complicated regions of contact between the hand and the object. Understanding and modeling this contact can potentially improve hand models, AR/VR experiences, and robotic grasping. Yet, we currently lack datasets of hand-object contact paired with other data modalities, which is crucial for developing and evaluating contact modeling techniques. We introduce ContactPose, the first dataset of hand-object contact paired with hand pose, object pose, and RGB-D images. ContactPose has 2306 unique grasps of 25 household objects grasped with 2 functional intents by 50 participants, and more than 2.9 M RGB-D grasp images. Analysis of ContactPose data reveals interesting relationships between hand pose and contact. We use this data to rigorously evaluate various data representations, heuristics from the literature, and learning methods for contact modeling. Data, code, and trained models are available at \url{https://contactpose.cc.gatech.edu}.
\end{abstract}

\section{Introduction} \label{sec:intro}
\begin{figure}
\centering
\includegraphics[width=\textwidth]{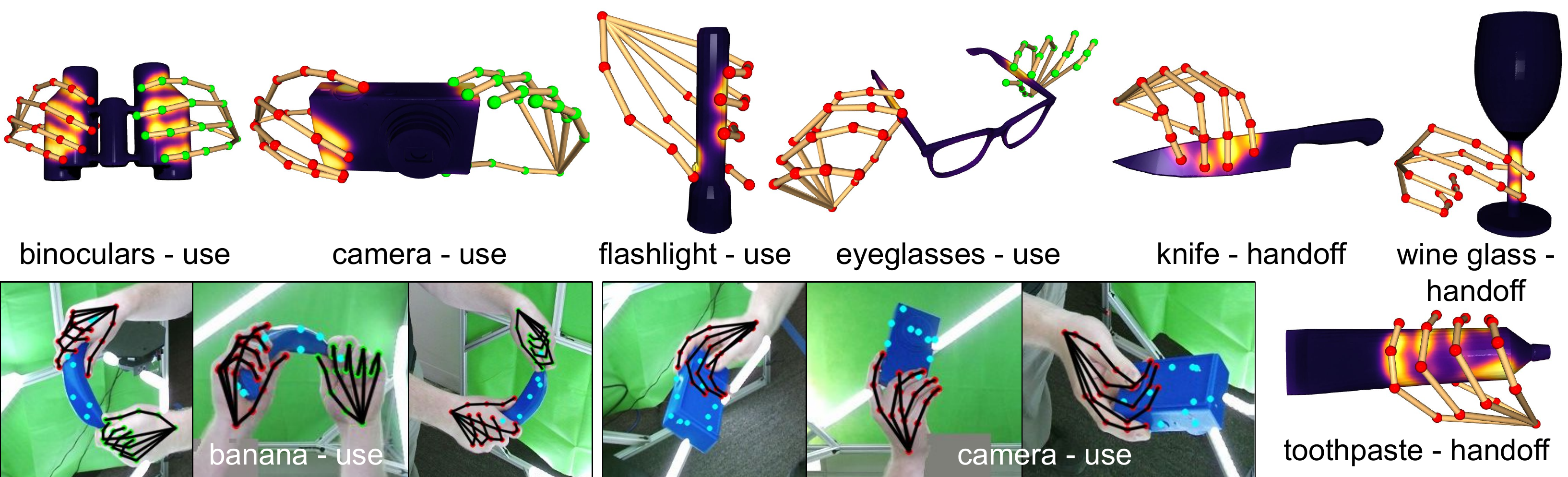}
\caption{Examples from ContactPose, a dataset capturing grasps of household objects. ContactPose includes high-resolution contact maps (object meshes textured with contact), 3D joints, and multi-view RGB-D videos of grasps. Left hand joints are \textbf{\textcolor{OliveGreen}{green}}, right hand joints are \textbf{\textcolor{red}{red}}.}
\label{fig:teaser}
\end{figure}
A person's daily experience includes numerous and varied hand-object interactions. Understanding and reconstructing hand-object interaction has received growing attention from the computer vision, computer graphics, and robotics communities. Most research has focused on hand pose estimation~\cite{NYUDataset,simon2017hand,FHAD_FirstPersonAction,tekin2019h+}, realistic hand and body reconstruction~\cite{handsInActionDataset,learningJointReconstructionOfHandsAndManipulatedObjects,Hassan_2019_ICCV,Zhang_2019_ICCV}, and robotic grasp prediction for anthropomorphic hands~\cite{brahmbhatt2019contactgrasp,lu2017planning}. In this paper, we address the under-explored problem of \textit{hand-object contact modeling} \ie predicting object contact with the hand, based on other information about the grasp, such as the 3D hand pose and grasp images. Accurate contact models have numerous applications in computer interfaces, understanding social interaction, object manipulation, and safety. For example, a hand contact model could interpret computer commands from physical interactions with a 3D printed replica object, or estimate if pathogens from a contaminated surface were transmitted through contact. More broadly, accurate contact modeling can improve estimation of grasp dynamics~\cite{ferrari1992planning,pollard1994parallel,graspit, mahler2019learning}, which can lead to better VR simulations of grasping scenarios and grasping with soft robotic hands~\cite{deimel2016novel,homberg2015haptic}.

Lack of ground-truth data has likely played a role in the under-exploration of this problem. Typically, the contacting surfaces of a grasp are occluded from direct observation with visible light imaging. Approaches that instrument the hand with gloves~\cite{wade2017force, tactileGloveNature} can subtly influence natural grasping behavior, and do not measure contact on the object surface. Approaches that intersect hand models with object models require careful selection of proximity thresholds or specific contact hand points \cite{learningJointReconstructionOfHandsAndManipulatedObjects, handsInActionDataset}. They also cannot account for soft hand tissue deformation, since existing state-of-the-art hand models~\cite{romero2017embodied} are rigid.

Brahmbhatt \etal~\cite{contactdbv1} recently introduced thermal cameras as sensors for capturing detailed ground-truth contact. Their method observes the heat transferred from the (warm) hand to the object through a thermal camera after the grasp. We adopt their method because it avoids the pitfalls mentioned above and allows for evaluation of contact modeling approaches with ground-truth data. However, it also imposes some constraints. 1) Objects have a plain visual texture since they are 3D printed to ensure consistent thermal properties. This does not affect 3D hand pose-based contact modeling methods and VR/robotic grasping simulators, since they rely on 3D shape and not texture. It does limit the generalization ability of RGB-based methods, which can potentially be mitigated by using depth images and synthetic textures. 2) The grasps are static, because in-hand manipulation results in multiple overlapping thermal hand-prints that depend on timing and other factors. Contact modeling for static grasps is still an unsolved problem, and forms the basis for future work on dynamic grasps. The methods we present here could be applied to dynamic scenarios frame-by-frame.

In addition, we develop a data collection protocol that captures multi-view RGB-D videos of the grasp, and an algorithm for 3D reconstruction of hand joints (\S~\ref{sec:data_protocol}). To summarize, we make the following contributions:
\begin{itemize}[noitemsep]
    \item \textbf{Data}: Our dataset (ContactPose) captures 50 participants grasping 25 objects with 2 functional intents. It includes high-quality contact maps for each grasp, over 2.9 M RGB-D images from 3 viewpoints, and object pose and 3D hand joints for each frame. We will make it publicly available to encourage research in hand-object interaction and pose estimation.
    \item \textbf{Analysis}: We dissect this data in various ways to explore the interesting relationship between contact and hand pose. This reveals some surprising patterns, and confirms some common intuitions.
    \item \textbf{Algorithms}: We explore various representations of object shape, hand pose, contact, and network architectures for learning-based contact modeling. Importantly, we rigorously evaluate these methods (and heuristic methods from the literature) against ground-truth unique to ContactPose.
\end{itemize}

\section{Related Work} \label{sec:related_work}

\begin{figure}
  \includegraphics[width=0.95\textwidth]{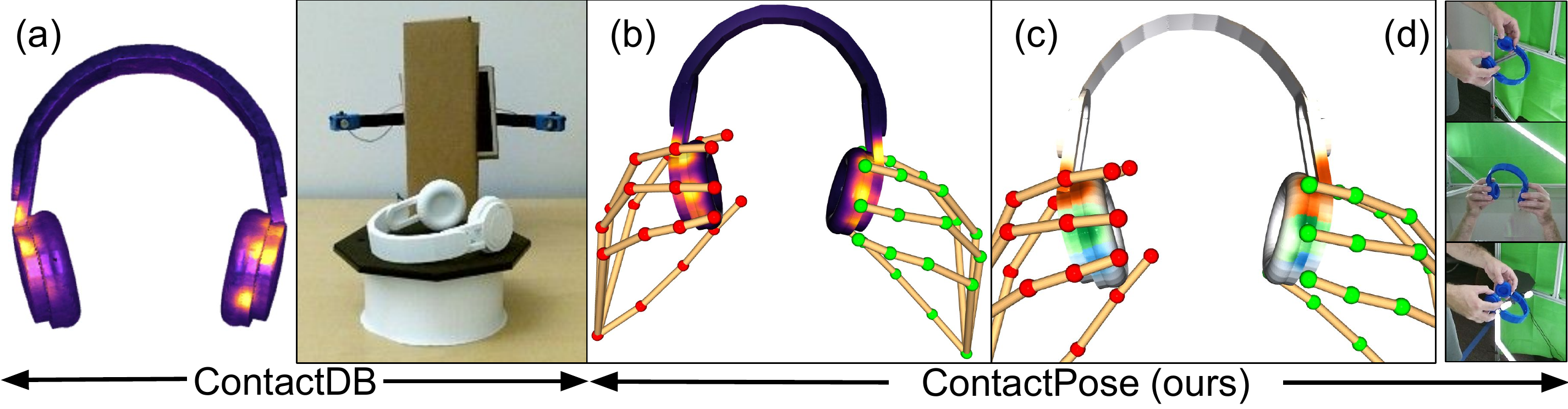}
  \caption{Comparison to ContactDB~\cite{contactdbv1}. It includes contact maps and turntable RGB-D images (a), which are often not enough to fully interpret the grasp e.g. it is not clear which fingers generated the contact. In contrast, ContactPose includes 3D joint locations (b), which allows association of contacted areas to hand parts (c), and multi-view RGB-D grasp images (d). These data enable a more comprehensive interpretation of the grasp.}
  \label{fig:contactdb_comparison}
\end{figure}

\newcommand{\y}{\textbf{\textcolor{OliveGreen}{\checkmark}}}
\newcommand{\n}{\textcolor{Red}{$\mathbf{\times}$}}
\begin{table}
\centering
\resizebox{\textwidth}{!}{
  \begin{tabular}{c|c|c|c|c|c|c|c}
    \textbf{Feature} & 
    \textbf{FPHA}~\cite{FHAD_FirstPersonAction} & 
    \textbf{HO-3D}~\cite{hampali2019ho} & 
    \textbf{FreiHand}~\cite{Freihand2019} & 
    \textbf{STAG}~\cite{tactileGloveNature} & 
    \textbf{ContactDB}~\cite{contactdbv1} &
    \textbf{Ours} \\
    \hline
    3D joints                           &\y &\y &\y &\n &\n &\y \\
    Object pose                       &\y &\y &\n &\n &\y &\y \\
    Grasp RGB images             &\y &\y &\y &\y &\n &\y \\
    Grasp Depth images          &\y &\y &\n &\n &\n &\y \\
    Natural hand appearance   &\n &\y &\y &\n &\n &\y\\
    Natural object appearance &\n &\y &\y &\y &\n &\n\\
    Naturally situated              &\y &\n &\n &\n &\n &\n\\
    Multi-view images             &\n &\n &\y &\n &\n &\y \\
    Functional intent               &\y &\n &\n &\n &\y &\y \\
    Hand-object contact         &\n &\n &\n &\y &\y &\y \\
    \# Participants                  &6  &8  &32 &1  &50 &50\\
    \# Objects                        &4  &8  &35 &26 &50 &25\\
    \end{tabular}}  
\caption{Comparison with existing hand-object datasets. ContactPose stands out for its size, and paired hand-object contact, hand pose and object pose.}
\label{tab:dataset_comparison}
\end{table}
\noindent\textbf{Capturing and modeling contact}: Previous works have instrumented hands and/or objects to capture contact. Bernardin \etal~\cite{bernardin2005sensor} and Sundaram \etal~\cite{tactileGloveNature} used a tactile glove to capture hand contact during grasping. Brahmbhatt \etal \cite{contactdbv1} used a thermal camera after the grasp to observe the heat residue left by the warm hand on the object surface. However, these datasets lacked either hand pose or grasp images, which are necessary for developing applicable contact models (Figure~\ref{fig:contactdb_comparison}). Pham \etal \cite{pham2015towards, Pham2018HandObjectCF} and Ehsani \etal \cite{Ehsani_2020_CVPR} tracked hands and objects in videos, and trained models to predict contact forces and locations at fingertips that explain observed object motion. In contrast, we focus on detailed contact modeling for complex objects and grasps, evaluated against contact maps over the entire object surface.

\noindent\textbf{Contact heuristics}: Heuristic methods to detect hand-object contact are often aimed at improving hand pose estimation. Hamer \etal \cite{hamer2010object} performed joint hand tracking and object reconstruction \cite{hamer2009tracking}, and inferred contact only at fingertips using proximity threshold. In simulation \cite{synthesisOfHandManipulationUsingContactSampling} and robotic grasping \cite{graspit,mahler2016dex}, contact is often determined similarly, or through collision detection \cite{teschner2005collision,larsen2000fast}. Ballan \etal \cite{ballan2012motion} defined a cone circumscribing object mesh triangles, and penalized penetrating hand points (and vice versa). This formulation has also been used to penalize self-penetration and environment collision~\cite{handsInActionDataset, SMPL-X}. While such methods were evaluated only through proxy tasks (\eg hand pose estimation), ContactPose enables evaluation against ground-truth contact (\S~\ref{sec:results}).

\noindent\textbf{Grasp Datasets}: Focusing on datasets involving hand-object interaction, hand pose has been captured in 3D with magnetic trackers \cite{FHAD_FirstPersonAction}, gloves \cite{glauser2019interactive,bernardin2005sensor}, optimization \cite{hampali2019ho}, multi-view boot-strapping \cite{simon2017hand}, semi-automated human-in-the-loop \cite{Freihand2019}, manually \cite{sridhar2016real}, synthetically \cite{learningJointReconstructionOfHandsAndManipulatedObjects}, or as instances of a taxonomy \cite{feix2015grasp, bullock2015yale,rogez2015understanding} along with RGB-D images depicting the grasps. However, none of these have contact annotations (see Table~\ref{tab:dataset_comparison}), and suffer additional drawbacks like lack of object information \cite{Freihand2019, simon2017hand} and simplistic objects \cite{sridhar2016real, FHAD_FirstPersonAction} and interactions \cite{sridhar2016real,learningJointReconstructionOfHandsAndManipulatedObjects}, which make them unsuitable for our task. In contrast, ContactPose has a large amount of ground-truth contact, and real RGB-D images of complex (including bi-manual) functional grasps for complex objects. The plain object texture is a drawback of ContactPose. Tradeoffs for this in the context of contact modeling are discussed in \S~\ref{sec:intro}.
\section{The ContactPose Dataset} \label{sec:dataset}
In ContactPose, hand-object contact is represented as a contact map on the object mesh surface, and observed through a thermal camera. Hand pose is represented as 3D hand(s) joint locations in the object frame, and observed through multi-view RGB-D video clips. The cameras are calibrated and object pose is known, so that the 3D joints can be projected into images (examples shown in supplementary material). Importantly, we avoid instrumenting the hands with data gloves, magnetic trackers or other sensors. This has the dual advantage of not interfering with natural grasping behavior and allowing us to use the thermal camera-based contact capture method from~\cite{contactdbv1}. We develop a computational approach (Section~\ref{sec:dataset_optim}) that optimizes for the 3D joint locations by leveraging accurate object tracking and aggregating over multi-view and temporal information. Our data collection protocol, described below, facilitates this approach.

\subsection{Data Capture Protocol and Equipment}\label{sec:data_protocol}

\begin{figure*}
  \begin{subfigure}[b]{0.24\textwidth}
    \includegraphics[width=\textwidth]{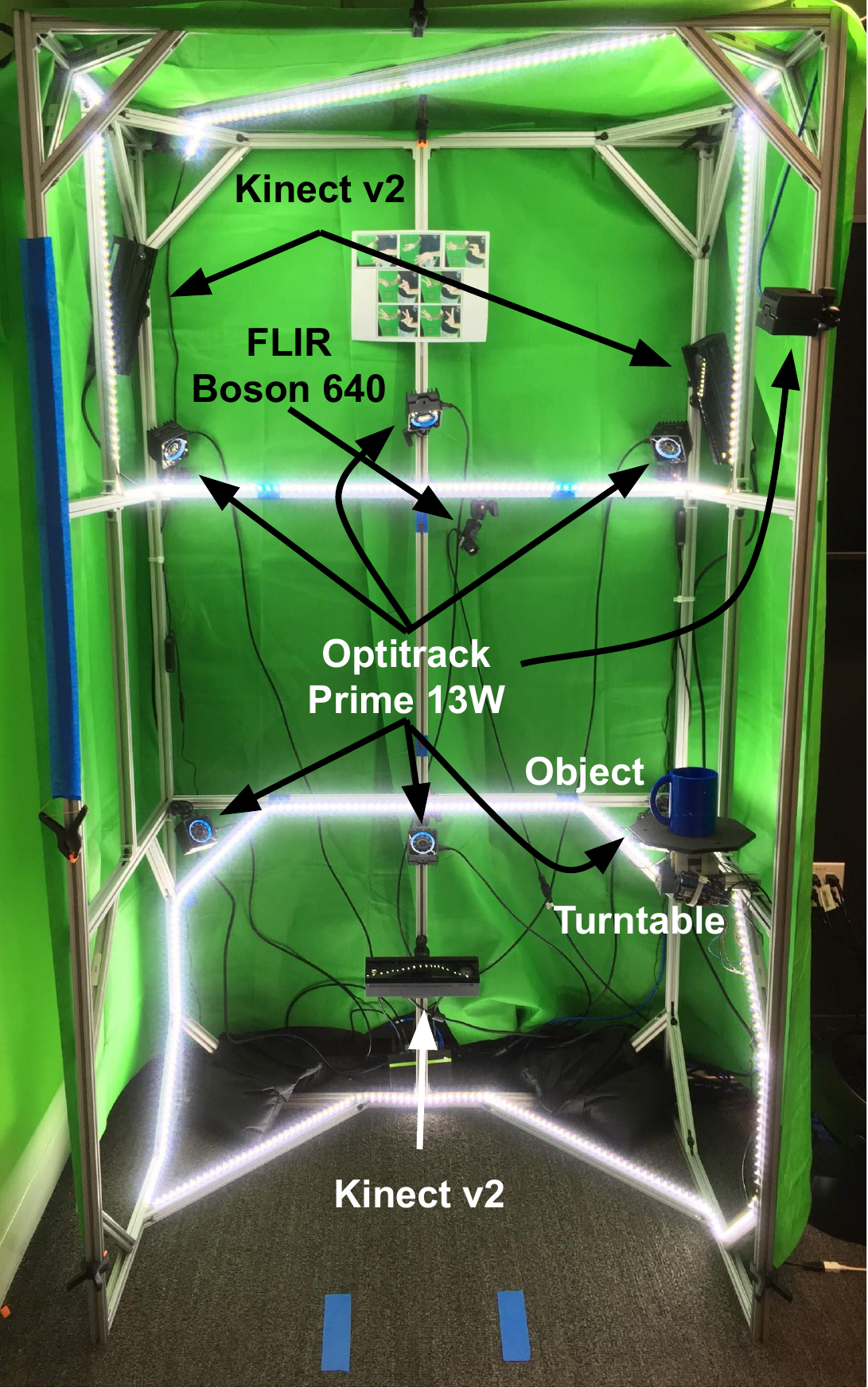}
    \caption{}
    \label{fig:capture_setup}
  \end{subfigure}
  \begin{subfigure}[b]{0.74\textwidth}
    \includegraphics[width=\textwidth]{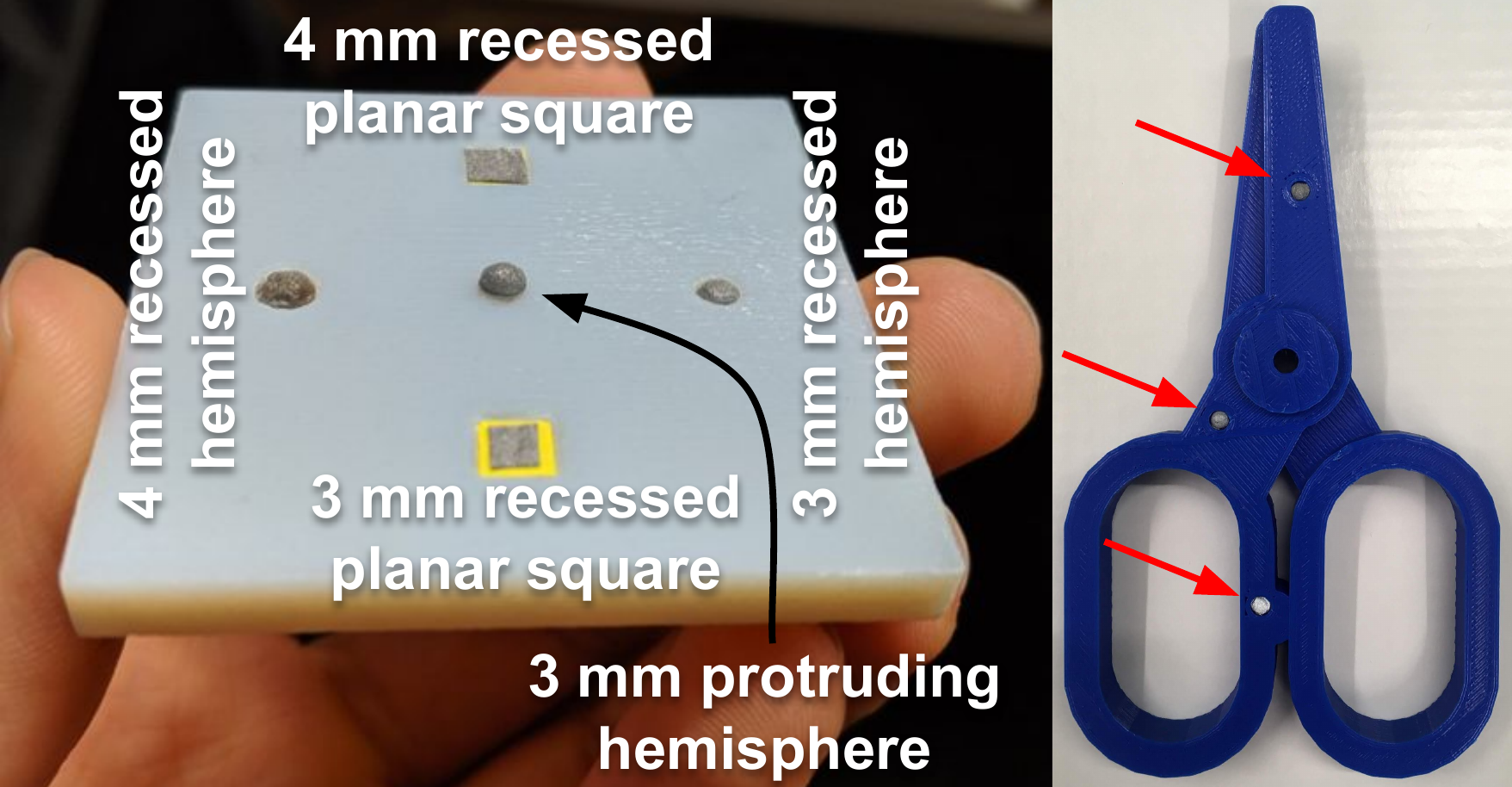}
    \caption{}
    \label{fig:marker_configs}
  \end{subfigure}
  \caption{(a) Our setup consists of 7 Optitrack Prime 13W tracking cameras, 3 Kinect v2 RGB-D cameras, a FLIR Boson 640 thermal camera, 3D printed objects, and a turntable. (b) \textbf{Left}: Different object tracking marker configurations we investigate. \textbf{Right}: 3D printed object with recessed 3 mm hemispherical markers (highlighted by \textbf{\textcolor{red}{red}} arrows) offer a good compromise between unobtrusiveness and tracking performance.}
\end{figure*}

We invite able-bodied participants to our laboratory and collect data through the following IRB-approved protocol. Objects are placed at random locations on a table in orientation normally encountered in practice. Participants are instructed to grasp an object with one of two functional intents (either using the object, or handing it off). Next, they stand in the data collection area (Figure~\ref{fig:capture_setup}) and move the object for 10-15 s in the cubical space. They are instructed to hold their hand joints steady, but are free to arbitrarily rotate the wrist and elbow, and to grasp objects with both hands or their dominant hand. This motion is recorded by 3 Kinect v2 RGB-D cameras (used for hand pose) and an Optitrack motion capture (mocap) system (used for object pose). Next, they hand the object to a researcher, who places it on a turntable, handling it with gloved hands. The object is recorded with the mocap system, Kinect v2, and a FLIR Boson 640 thermal camera as the turntable rotates a circle.

\noindent\textbf{Contact Capture}: Thermal images are texture-mapped to the object mesh using Open3D \cite{Zhou2018open3d, zhou2014color}. As shown in \cite{contactdbv1} and the supp. mat., the resulting mesh textures (called contact maps) accurately capture hand-object contact.

\noindent\textbf{Object Selection and Fabrication}: We capture grasps on a subset of 25 objects from~\cite{contactdbv1} that are applicable for both `use' and `hand-off' grasping (see supp. mat. for a list). The objects are 3D printed in blue for good contrast with hands and the green background of our capture area. 3D printing the objects ensures consistent thermal properties and ensures geometric consistency between real world objects in capture sessions and the 3D models in our dataset.

Mocap recovers the object pose using retro-reflective markers, whose the placement on the object requires some care.  Attaching a large `marker tree' would block interactions with a significant area of the surface.  Placing hemispherical markers on the surface is more promising, but a sufficient number (8+) are needed to ensure visibility during hand occlusion and the resulting `bumps' can be uncomfortable to touch, which might influence natural grasping behavior. We investigate a few alternative marker configurations (Figure~\ref{fig:marker_configs}).  Flat pieces of tape were more comfortable but only tracked well when the marker was directly facing the camera.  A good compromise is to use 3 mm hemispherical markers but to recess them into the surface by adding small cut-outs during 3D printing. These are visible from a wide range of angles but do not significantly affect the user's grip. Fixing the marker locations also allows for simple calibration between the Optitrack rigid body and the object's frame.

\subsection{Grasp Capture without Hand Markers}\label{sec:dataset_optim}
Each grasp is observed through $N$ frames of RGB-D images from $C$ cameras. We assume that the hand is fixed relative to the object, and the 6-DOF object pose for each frame is given. So instead of estimating 3D joints separately in each frame, we can aggregate the noisy per-frame 2D joint detections into a single set of high-quality 3D joints, which can be transformed by the frame's object pose.

For each RGB frame, we use Detectron \cite{maskRCNN} to locate the wrist, and run the OpenPose hand keypoint detector~\cite{simon2017hand} on a 200$\times$200 crop around the wrist. This produces 2D joint detections $\lbrace \mathbf{x}^{(i)} \rbrace_{i=1}^N$ and confidence values $\lbrace \mathbf{w}^{(i)} \rbrace_{i=1}^N$, following the 21-joint format from~\cite{simon2017hand}.  One option is to lift these 2D joint locations to 3D using the depth image~\cite{NYUDataset}, but that biases the location toward the camera and the hand surface (our goal is to estimate joint locations internal to the hand). Furthermore, the joint detections at any given frame are unreliable.  Instead, we use our hand-object rigidity assumption to estimate the 3D joint locations $\mathbf{^oX}$ in the object frame that are consistent with all $NC$ images. This is done by minimizing the average re-projection error:
\begin{equation} \label{eq:optim}
\min_{\mathbf{^oX}} \sum_{i=1}^N \sum_{c=1}^C \mathcal{D} \left( \mathbf{x}^{(i)}_c, \pi \left( \mathbf{^oX}; K_c, ^cT_w~^wT_o^{(i)} \right); \mathbf{w}^{(i)}_c \right)
\end{equation}
where $\mathcal{D}$ is a distance function, and $\pi(\cdot)$ is the camera projection function using camera intrinsics $K_c$ and object pose w.r.t. camera at frame $i$, $^cT_o^{(i)} =~^cT_w~^wT_o^{(i)}$. Our approach requires the object pose w.r.t. world at each frame $^wT_o^{(i)}$ i.e. object tracking. This is done using an Optitrack motion capture system tracking markers embedded in the object surface.

In practice, the 2D joint detections are noisy and object tracking fails in some frames. We mitigate this by using the robust Huber function~\cite{huber1992robust} over Mahalanobis distance ($\mathbf{w}^{(i)}$ acting as variance) as $\mathcal{D}$, and wrapping Eq.~\ref{eq:optim} in a RANSAC~\cite{fischler1981random} loop. A second pass targets frames that fail the RANSAC inlier test due to inaccurate object pose. Their object pose is estimated through the correspondence between their 2D detections and the RANSAC-fit 3D joint locations, and they are included in the inlier set if they pass the inlier test (re-projection error less than a threshold). It is straightforward to extend the optimization described above to bi-manual grasps. We manually curated the dataset, including clicking 2D joint locations to aid the 3D reconstruction in some cases, and discarding some obviously noisy data. 

\begin{figure*}
 \includegraphics[width=\textwidth]{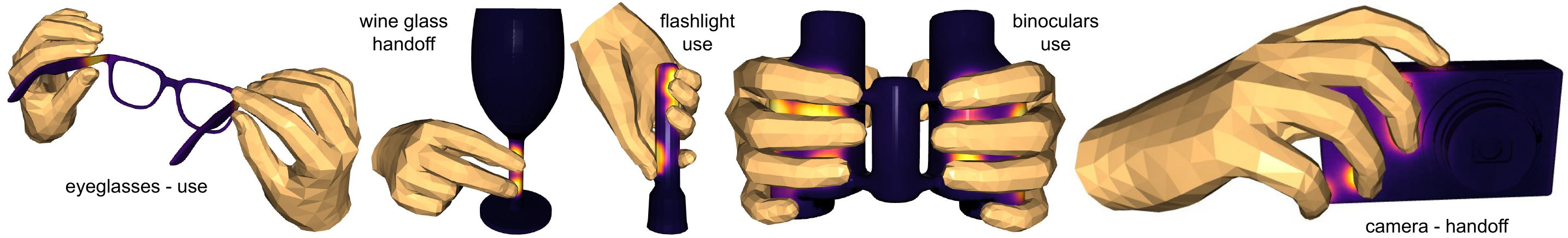}
 \caption{MANO hand meshes~\cite{romero2017embodied} fit to ContactPose data. Both hand pose and shape parameters are optimized to minimize the distance of MANO joints from ContactPose 3D joint annotations.}
 \label{fig:mano_fits}
\end{figure*}

\noindent\textbf{Hand Mesh Models}: In addition to capturing grasps, hand shape information is collected through palm contact maps on a flat plate, and multi-view RGB-D videos of the participant performing 7 known hand gestures (shown in the supplementary material). Along with 3D joints, this data can potentially enable fitting of the MANO hand mesh model~\cite{romero2017embodied} to each grasp \cite{moon2018v2v}. In this paper, we use meshes fit to 3D joints (Figure~\ref{fig:mano_fits}, see supp. mat. for details) for some of the analysis and learning experiments discussed below.

\section{Data Analysis} \label{sec:analysis}
Contact maps are $[0, 1]$ normalized following the sigmoid fitting procedure from~\cite{contactdbv1}.

\noindent\textbf{Association of Contact to Hand Parts}: It has been observed that certain fingers and parts (e.g. fingertips) are contacted more frequently than others~\cite{bullock2015yale, bullock2013grasp}. ContactPose allows us to quantify this. This can potentially inform anthropomorphic robotic hand design and tactile sensor (e.g. BioTac~\cite{biotac}) placement in robotic hands. For each grasp, we threshold the contact map at 0.4 and associate each contacted object point with its nearest hand point from the fitted MANO hand mesh. A hand point is considered to be contacted if one or more contacted object points are associated with it. A coarser analysis at the phalange level is possible by modeling phalanges as line segments connecting joints. In this case, the distance from an object point to a phalange is the distance to the closest point on the line segment.

\begin{figure*}
  \begin{subfigure}[b]{0.70\textwidth}
    \includegraphics[width=\textwidth]{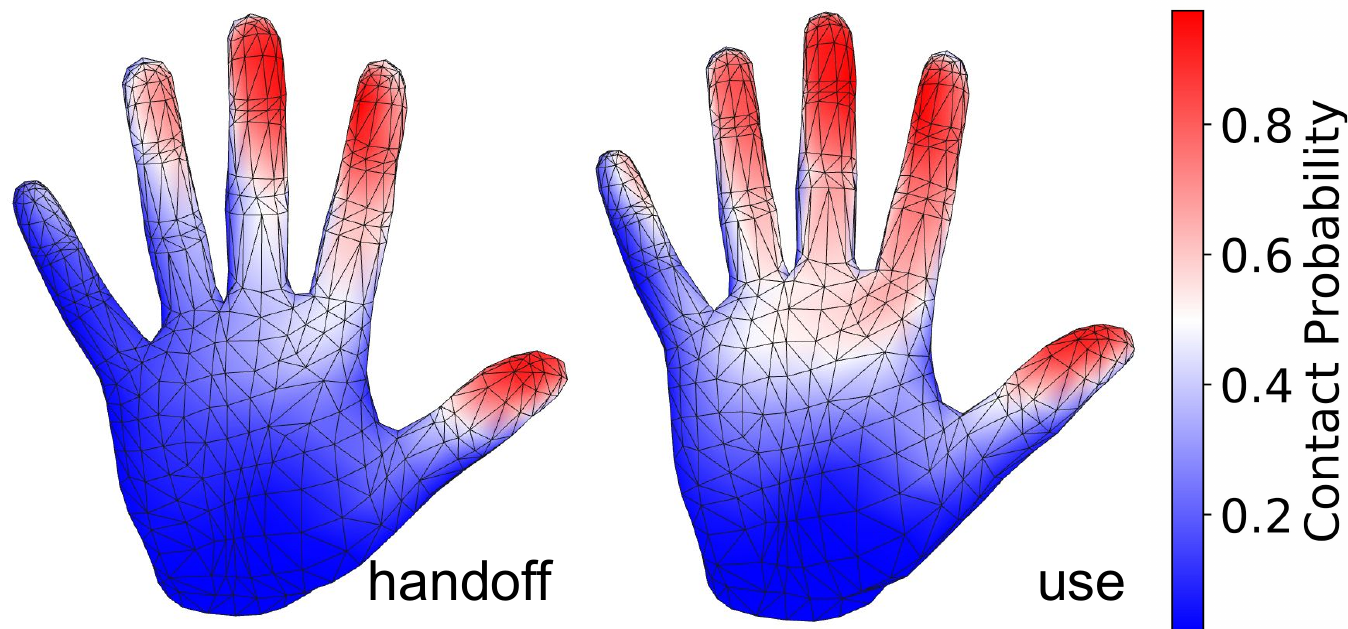}
    \caption{}
    \label{fig:hand_contact_prob}
  \end{subfigure}
  \begin{subfigure}[b]{0.29\textwidth}
    \includegraphics[width=\textwidth]{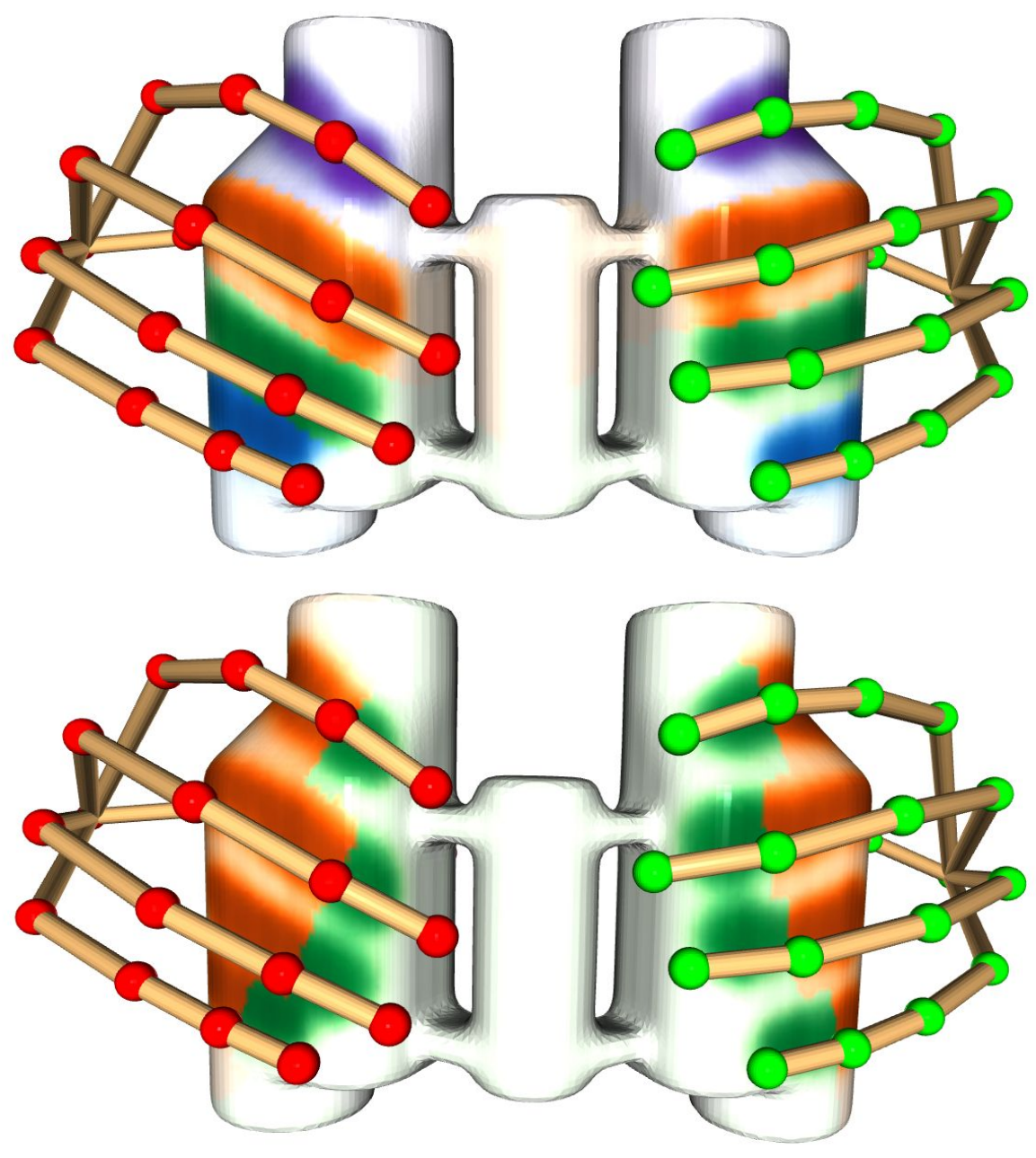}
    \caption{}
    \label{fig:semantic_contact}
  \end{subfigure}
  \caption{(a) Hand contact probabilities estimated from the entire dataset. (b) Association of contacted binoculars points with fingers (top) and sets of phalanges at the same level of wrist proximity (bottom), \textbf{\textcolor{RoyalPurple}{indicated}} \textbf{\textcolor{Orange}{by}} \textbf{\textcolor{OliveGreen}{different}} \textbf{\textcolor{ProcessBlue}{colors}}.}
\end{figure*}

Figure~\ref{fig:hand_contact_prob} shows the contact probabilities averaged over `use' and `hand-off' grasps. Not surprisingly, the thumb, index, and middle finger are the most contacted fingers, and tips are the most contacted phalanges. Even though fingertips receive much attention in grasping literature, the contact probability for all three phalanges of the index finger is \emph{higher} than that of the pinky fingertip. Proximal phalanges and palm also have significant contact probabilities. This is consistent with observations made by Brahmbhatt et al~\cite{contactdbv1}. Interestingly, contact is more concentrated at the thumb and index finger for `hand-off' than `use'. `Use' grasps have an average contact area of 35.87 cm$^2$ compared to 30.58 cm$^2$ for `hand-off'. This analysis is similar to that in Fig. 3 of Hasson \etal \cite{learningJointReconstructionOfHandsAndManipulatedObjects}, but supported by ground-truth contact rather than synthetic grasps.

\begin{figure*}
  \begin{subfigure}{.51\textwidth}
    \includegraphics[width=\textwidth]{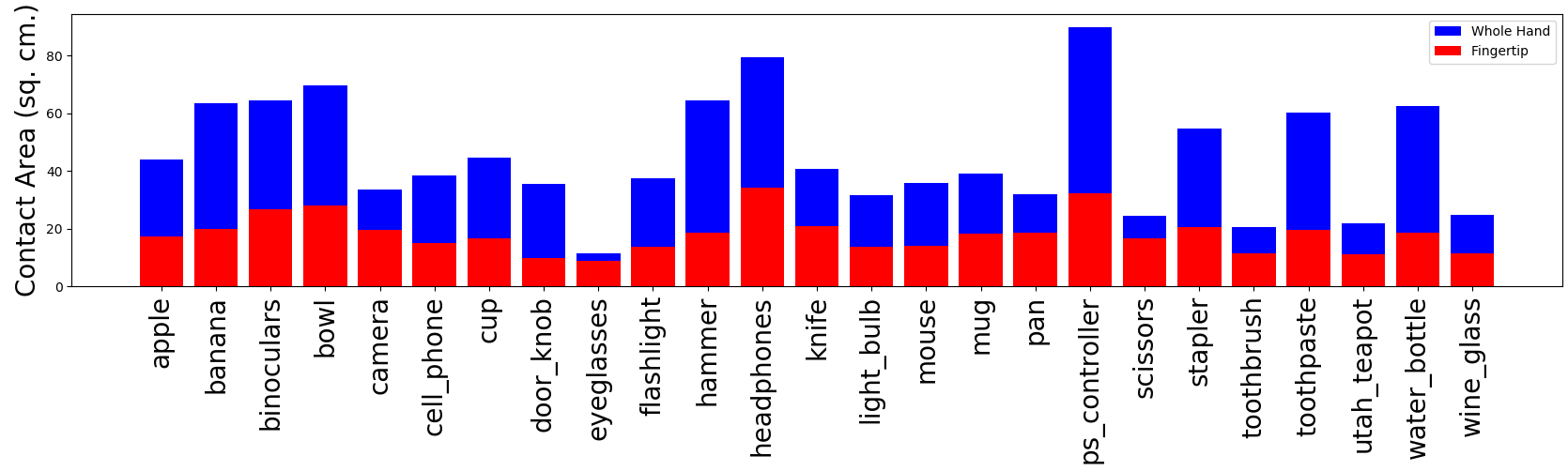}
    \caption{`use' grasps}
  \end{subfigure}
  \hfill
  \begin{subfigure}{.48\textwidth}
    \includegraphics[width=\textwidth]{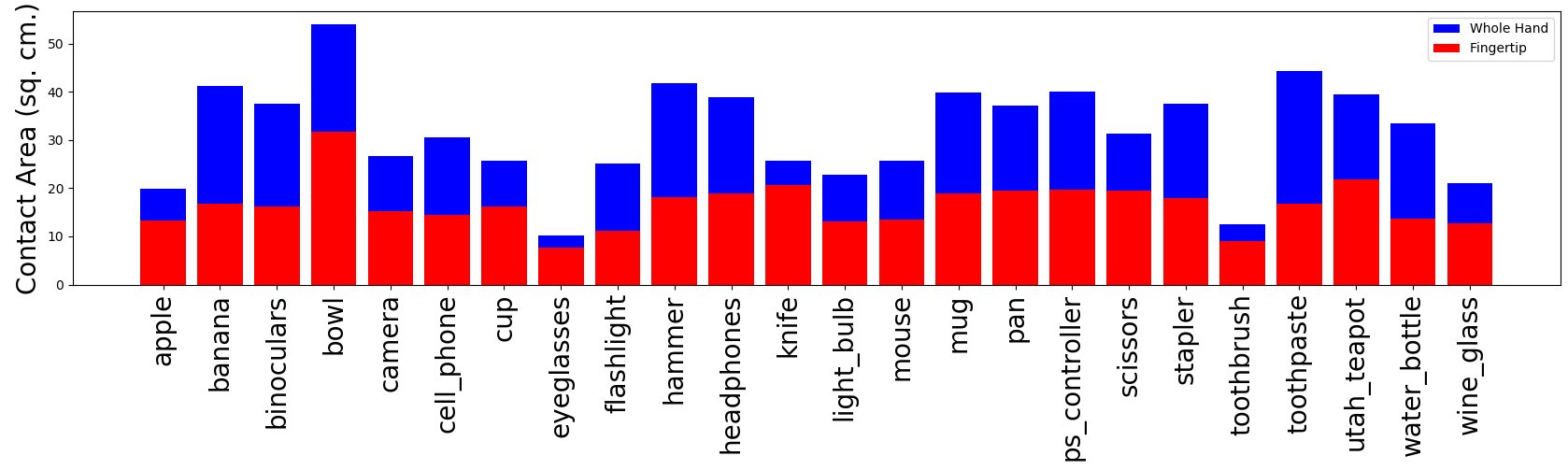}
    \caption{`handoff' grasps}
  \end{subfigure}
  \caption{Comparing average fingertip (\textbf{\textcolor{red}{red}}) vs. whole-hand (\textbf{\textcolor{blue}{blue}}) contact areas.}
  \label{fig:contact_areas}
\end{figure*}

Comparison of the average fingertip vs. whole-hand contact areas (Figure \ref{fig:contact_areas}) shows that non-fingertip areas play a significant role in grasp contact, confirming the approximate analysis in \cite{contactdbv1}.

\begin{figure}
  \includegraphics[width=\columnwidth]{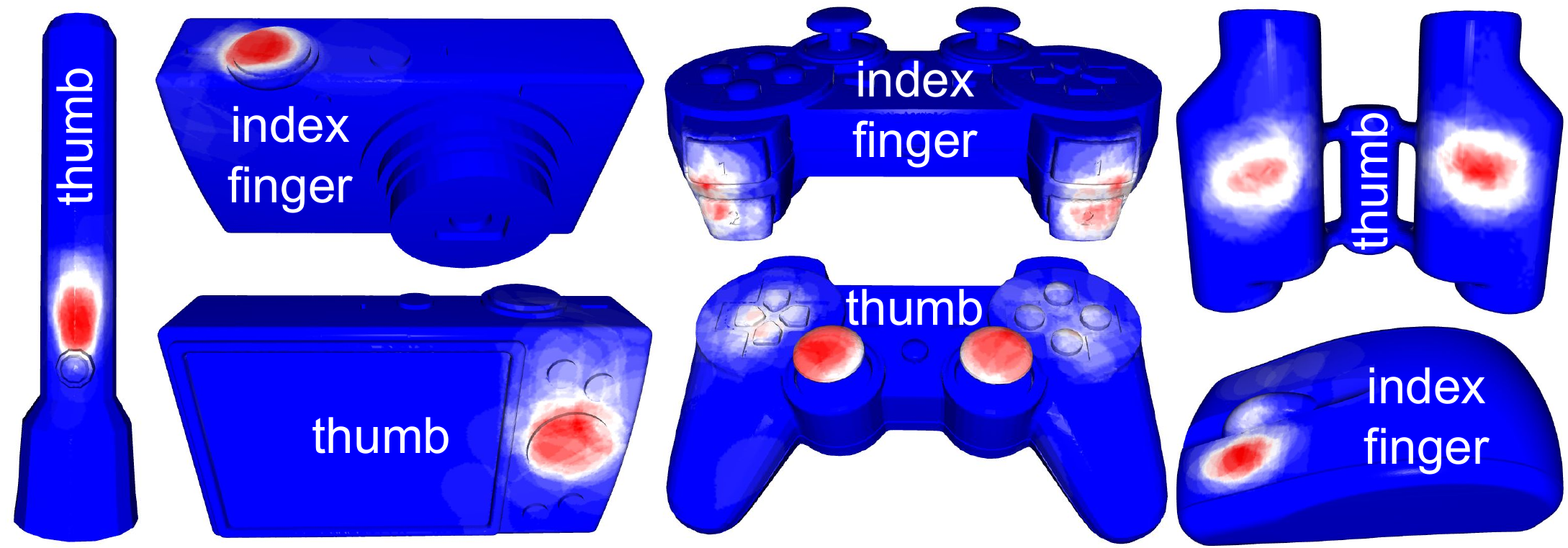}
  \caption{Automatic `active area' discovery: Contact probability for various hand parts on the object surface.}
  \label{fig:active_areas}
\end{figure}

\noindent\textbf{Automatic Active Area Discovery}: Brahmbhatt et al~\cite{contactdbv1} define active areas as regions on the object highly likely to be contacted. While they manually selected active areas and measured their probability of being contacted by \emph{any} part of the hand, ContactPose allows us to `discover' active areas automatically and for \emph{specific} hand parts. We use the object point-phalange association described above (\eg Fig. \ref{fig:semantic_contact}) to estimate the probability of each object point being contacted by a given hand part (e.g. index finger tip), which can be thresholded to segment the active areas. Figure~\ref{fig:active_areas} shows this probability for the index fingertip and thumb, for `use' grasps of some objects. This could potentially inform locations for placing contact sensors (real \cite{Pham2018HandObjectCF} or virtual for VR) on objects. 

\begin{figure*}
\begin{subfigure}{.61\textwidth}
	\includegraphics[width=\textwidth]{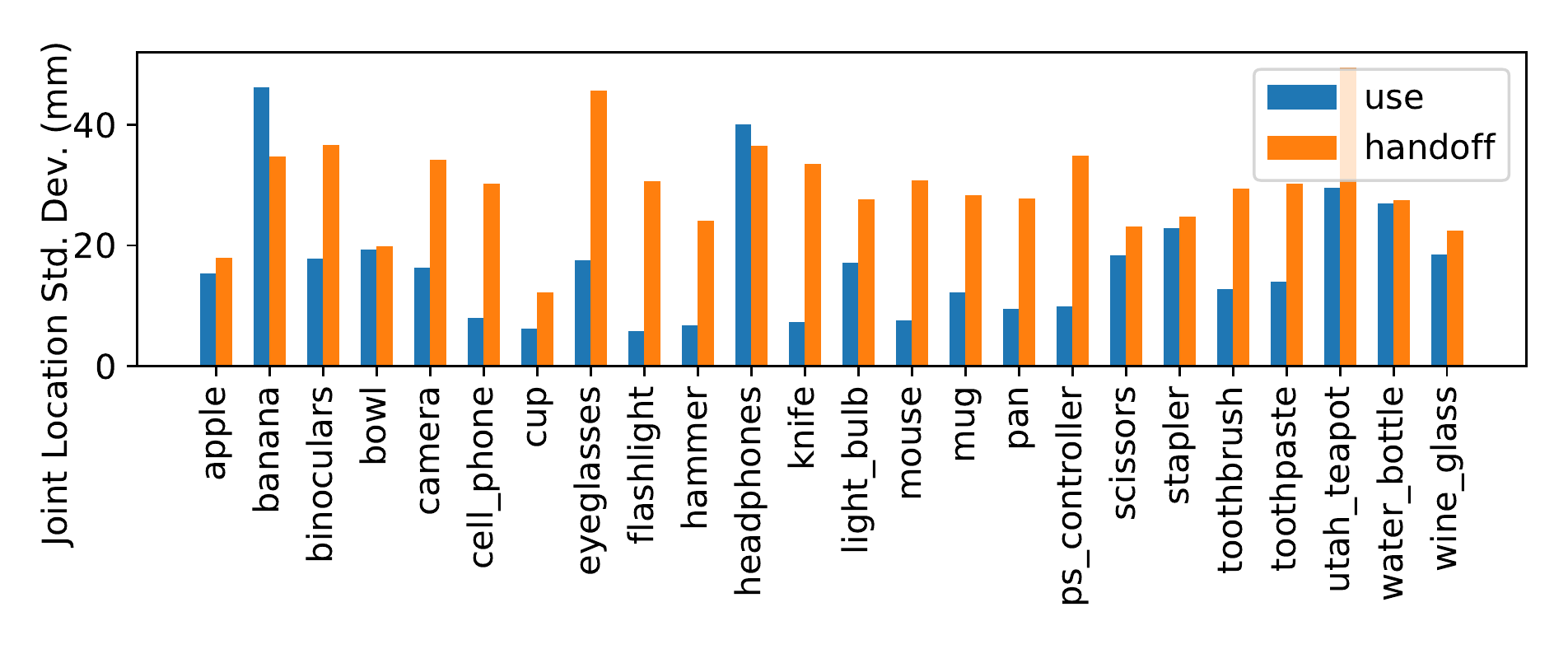}
	\caption{}
	\label{fig:joint_stdev}
\end{subfigure}
\hfill
\begin{subfigure}{.36\textwidth}
	\includegraphics[width=\textwidth]{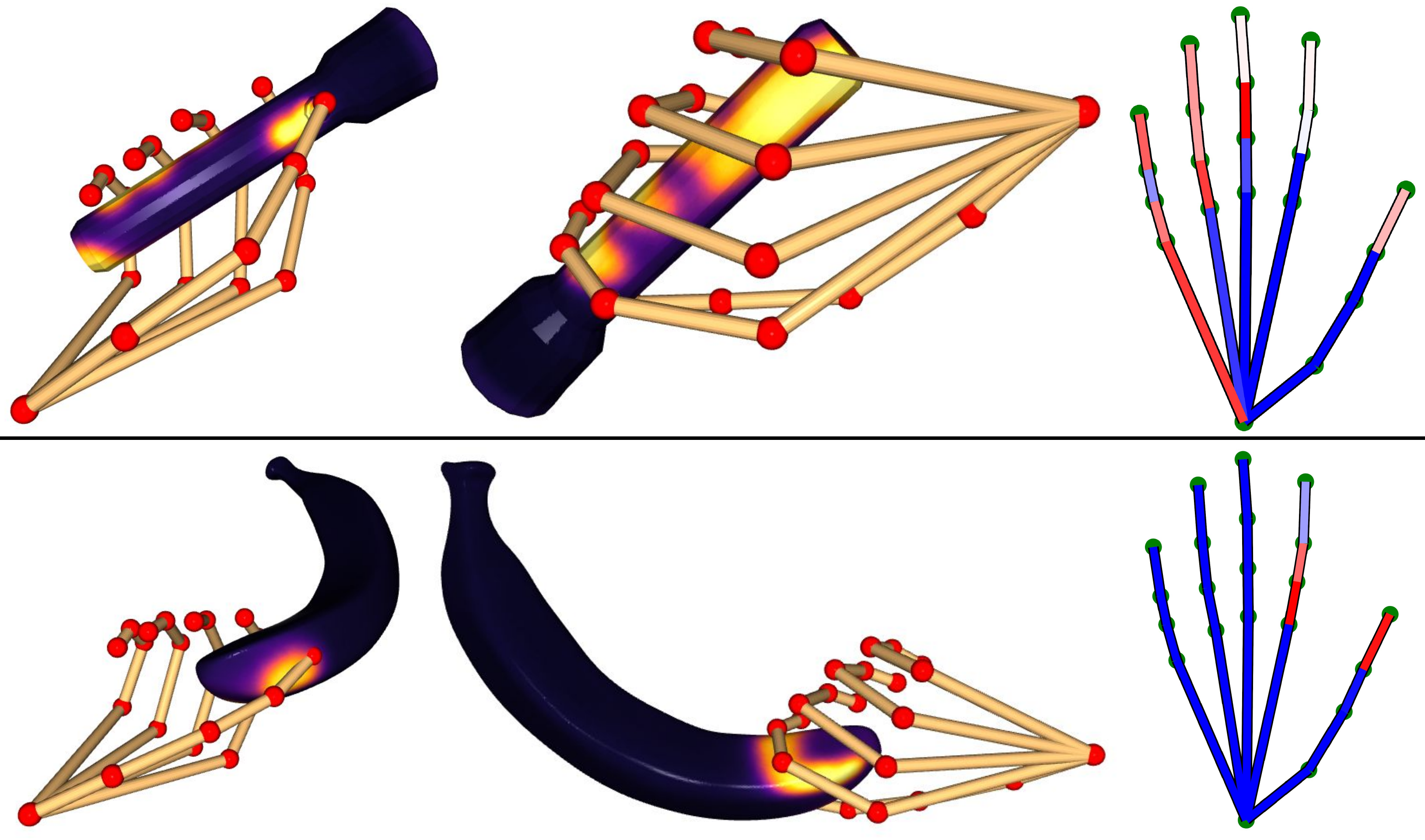}
	\caption{}
	\label{fig:pose_vs_contact_diversity}
\end{subfigure}
\caption{(a) Per-object standard deviation in 3D joint locations, for `use' and `hand-off'. `Hand-off' grasps consistently exhibit more diversity than `use' grasps. (b) A pair of grasps with similar hand pose but different contact characteristics. Hand contact feature color-coding is similar to Figure~\ref{fig:hand_contact_prob}.}
\end{figure*}

\begin{figure}
  \includegraphics[width=\textwidth]{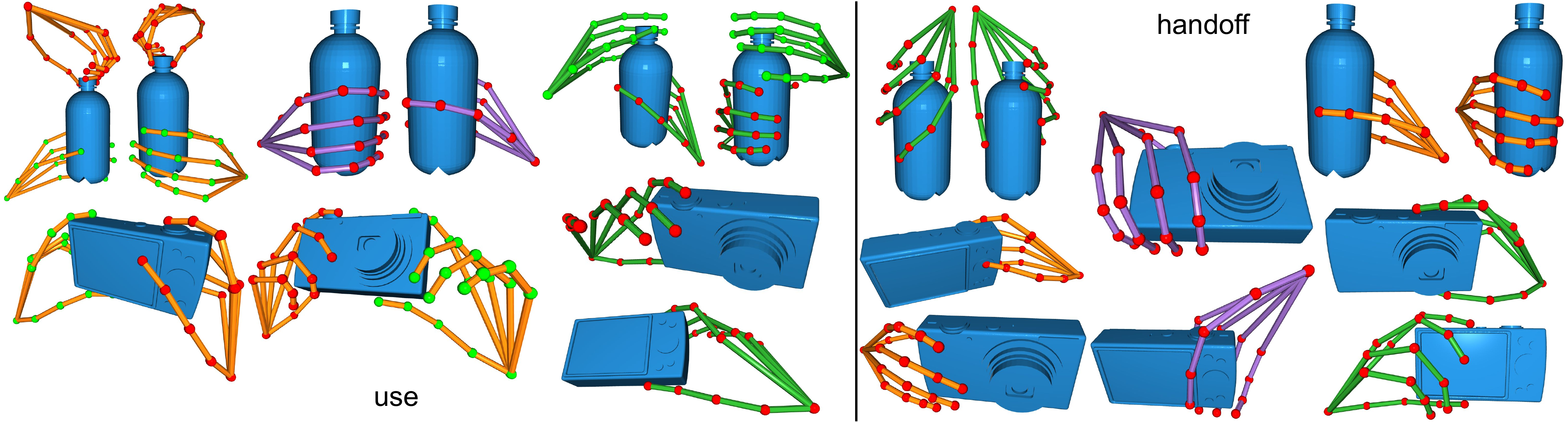}
  \caption{Examples from hand pose clusters for `use' and `hand-off' grasps. Grasps from different clusters are shown \textbf{\textcolor{Orange}{with}} \textbf{\textcolor{Green}{different}} \textbf{\textcolor{RoyalPurple}{colors}} (some grasps are bi-manual). Left hand joints are \textbf{\textcolor{OliveGreen}{green}}, right hand joints are \textbf{\textcolor{red}{red}}.}
  \label{fig:intra_object_clustering}
\end{figure}

\noindent\textbf{Grasp Diversity}: We further quantify the effect of intent on grasping behavior by measuring the standard deviation of 3D joint locations over the dataset. The mean of all 21 joint standard deviations is shown in Figure~\ref{fig:joint_stdev}. It shows that `hand-off' grasps are more diverse than `use' grasps in terms of hand pose. We accounted for symmetrical objects (e.g. wine glass) by aligning the 6 palm joints (wrist + 5 knuckles) of all hand poses for that object to a single set of palm joints, where the only degree of freedom for alignment is rotation around the symmetry axis. Hand size is normalized by scaling all joint location such that the distance from wrist to middle knuckle is constant.

Organizing the grasps by clustering these aligned 3D joints (using L2 distance and HDBSCAN~\cite{hdbscan}) reveals the diversity of grasps captured in ContactPose (Figure~\ref{fig:intra_object_clustering}). `Hand-off' grasps exhibit a more continuous variation than `use' grasps, which are tied more closely to the function of the object. The average intra-cluster distance for `use' grasps is 32.5\% less than that for `handoff' grasps.

Figure~\ref{fig:pose_vs_contact_diversity} shows pair of grasps found by minimizing hand pose distance and maximizing hand contact distance. We use the phalange-level contact association described above. Summing the areas of all object mesh triangles incident to all vertices associated with a phalange creates a 20-dimensional vector. We use L2 distance over this vector as contact distance. It shows that grasps with similar hand pose can contact different parts of the object and/or hand, inducing different forces and manipulation possibilities~\cite{FHAD_FirstPersonAction} and emphasizing that hand pose alone provides an inadequate representation of grasping.

\section{Contact Modeling Experiments} \label{sec:learning}
This section describes our experiments on \textit{contact modeling} given the hand pose or RGB grasp image(s), assuming known object geometry and pose. Our experiments focus on finding good data representations and learning algorithms, and evaluating techniques against ground-truth. By providing high-quality contact output from readily available input modalities, such models can enable better hand-object dynamics simulation in AR/VR and soft robotic grasping.

\noindent\textbf{Object Shape Representation}: We represent the object shape through either a pointcloud densely sampled from the surface (1K-30K points based on size), or a $64^3$ voxel occupancy grid. Features encoding the input hand pose are associated with individual points (voxels). The entire pointcloud (voxel grid) is then processed to predict contact values for points (surface voxels).

\noindent\textbf{Hand Pose Representation}: Features relating object shape to hand pose are computed for each point or voxel. These features have varying levels of richness of hand shape encoding. To simulate occlusion and noisy pose perception for the first 4 features, we sample a random camera pose and drop (set to 0) all features associated with the farthest 15\% of the joints from the camera.
\begin{itemize}
 \item \texttt{\textbf{simple-joints}}: We start by simply using the 21 3D joint locations w.r.t. the object coordinate system as 63-dimensional features for every point. For bi-manual grasps, points use the hand with the closest joint.
 \item \texttt{\textbf{relative-joints}}: Since contact at an object surface point depends on the \textit{relative} position of the finger, we next calculate relative vectors from an object point to every joint of the hand closest to it. Contact also depends on the surface geometry: a finger is more likely to contact an object point if the vector to it is parallel to the surface normal at that point. Hence we use unit-norm surface normals and the relative joint vectors to form $63+3=66$-dimensional features for every point.
 \item \texttt{\textbf{skeleton}}: To better capture hand joint connectivity, we compute relative vectors from an object point to the nearest point on phalanges, modeled as line segments. 40-dimensional features for each object point are constructed by concatenating the lengths of 20 such vectors (one for each phalange), and their dot product with the surface normal at that object point.
 \item \texttt{\textbf{mesh}}: These features leverage the rich MANO hand model geometry. A relative vector is constructed from the object point to its closest hand mesh point. 23-dimensional features are constructed from the length of this vector, its dot product with the surface normal, and distances to 21 hand joints.
 \item Grasp Image(s): To investigate if CNNs can extract relevant information directly from images, we extract dense 40-dimensional features from 256$\times$256 crops of RGB grasp images using a CNN encoder-decoder inspired by U-Net~\cite{ronneberger2015unet} (see supplementary material for architecture). These images come from the same time instant. We investigate both 3-view and 1-view settings, with feature extractor being shared across views for the former. Features are transferred to corresponding 3D object points using the known object pose and camera intrinsics, averaging the features if multiple images observe the same 3D point (Figure~\ref{fig:image_pred_architecture}). Points not visible from any image have all features set to 0. Image backgrounds are segmented by depth thresholding at the 20th percentile, and the foreground pixels are composited onto a random COCO~\cite{lin2014microsoft} image. This investigation is complementary to recent work on image-based estimation of object geometry \cite{zhou20153d,groueix2018papier}, object pose \cite{garon2017deep,tremblay2018deep}, and hand pose \cite{simon2017hand, Zhang_2019_ICCV, tekin2019h+, Freihand2019,hampali2019ho}.
\end{itemize}

\noindent\textbf{Contact Representation}: We observed in early experiments that the mean squared error loss resulted in blurred and saturated contact predictions. This might be due to contact value occurrence imbalance and discontinuous contact boundaries for smooth input features. Hence, we discretize the $[0, 1]$ normalized values into 10 equal bins and treat contact prediction as a classification problem, inspired by Zhang et al~\cite{zhang2016colorful}. We use the weighted cross entropy loss, where the weight for each bin is proportional to a linear combination of the inverse occurrence frequency of that bin and a uniform distribution (Eq. 4 from~\cite{zhang2016colorful} with $\lambda=0.4$). Following~\cite{zhang2016colorful}, we derive a point estimate for contact in $[0, 1]$ from classification outputs using the annealed mean ($T=0.1$).

\noindent\textbf{Learning Algorithms}: Given the hand pose features associated with points or voxels, the entire pointcloud or voxel grid is processed by a neural network to predict the contact map. We use the PointNet++~\cite{qi2017pointnet++} architecture implemented in pytorch-geometric~\cite{paszke2017automatic,pytorch_geometric} (modified to reduce the number of learnable parameters) for pointclouds, and the VoxNet~\cite{maturana2015voxnet}-inspired 3D CNN architecture from~\cite{contactdbv1} for voxel grids (see the supplementary material for architectures). For voxel grids, a binary feature indicating voxel occupancy is appended to hand pose features. Following \cite{contactdbv1}, hand pose features are set to 0 for voxels inside the object. Because the features are rich and provide fairly direct evidence of contact, we include a simple learner baseline of a multi-layer perceptron (MLP) with 90 hidden nodes, parametric ReLU~\cite{he2015delving} and batchnorm~\cite{ioffe2015batch}.

\noindent\textbf{Contact Modeling Heuristics}: We also investigate the effectiveness of heuristic techniques, given detailed hand geometry through the MANO hand mesh. Specifically, we use the conic distance field $\Psi$ from \cite{ballan2012motion,handsInActionDataset} as a proxy for contact intensity. To account for imperfections in hand modelling (due to rigidity of the MANO mesh) and fitting, we compute $\Psi$ not only for collisions, but also when the hand and object meshes are closer than 1 cm. Finally, we calibrate $\Psi$ to our ground truth contact through least-squares linear regression on 4700 randomly sampled contact points. Both these steps improve the technique's performance.  

\section{Results} \label{sec:results}

\begin{table}
\centering
\resizebox{\textwidth}{!}{
\begin{tabular}{c|c||c|c||c|c}

\multirow{2}{*}{\textbf{Learner}} & \multirow{2}{*}{\textbf{Features}} &
\multicolumn{2}{c||}{\textbf{Participant Split}} &
\multicolumn{2}{c}{\textbf{Object Split}} \\\cline{3-6}

& & \textbf{AuC (\%)} & \textbf{Rank} &
\textbf{AuC (\%)} & \textbf{Rank}\\\hline

None & Heuristic \cite{ballan2012motion, handsInActionDataset} &
78.31 & 5 & 81.11 & 4\\\hline

VoxNet~\cite{maturana2015voxnet, contactdbv1} & \textbf{\texttt{skeleton}} &
77.94 & & 79.99 &\\\hline

\multirow{4}{*}{MLP} & \textbf{\texttt{simple-joints}} &
75.11 & & 77.83 &\\

& \textbf{\texttt{relative-joints}} & 75.39 & & 78.83 &\\ 

& \textbf{\texttt{skeleton}} & 80.78 & 3 & 80.07 &\\ 

& \textbf{\texttt{mesh}} & 79.89 & 4 & \textbf{84.74} & 1\\\hline

\multirow{4}{*}{PointNet++} & \textbf{\texttt{simple-joints}} &
71.61 & & 73.67 &\\

& \textbf{\texttt{relative-joints}} & 74.51 & & 77.10 &\\ 

& \textbf{\texttt{skeleton}} & 81.15 & 2 & 81.49 & 3\\ 

& \textbf{\texttt{mesh}} & \textbf{81.29} & 1 & 84.18 & 2\\\hline

Image enc-dec, & images (1-view) & 72.89 & & 77.09 &\\\cline{2-6}

PointNet++ & images (3-view) & 78.06 & & 80.80 & 5\\
\end{tabular}}
\caption{Contact prediction re-balanced AuC (\%) (higher is better) for various combinations of features and learning methods.}
\label{tab:auc_numbers}
\end{table}

In this section, we evaluate various combinations of features and learning algorithms described in \S~\ref{sec:learning}. The metric for quantitative evaluation is the area under the curve formed by calculating accuracy at increasing contact difference thresholds. Following~\cite{zhang2016colorful}, this value is re-balanced to account for varying occurrence frequencies of values in the 10 contact bins. We create two data splits: the \emph{object split} holds out mug, pan and wine glass following \cite{contactdbv1}, and the \emph{participant split} holds out participants 5, 15, 25, 35, and 45. The held out data is used for evaluation, and models are trained on the rest.

Table \ref{tab:auc_numbers} shows the re-balanced AuC values averaged over held out data for the two splits. We observe that features capturing richer hand shape information perform better (\eg \texttt{\textbf{simple}}-\texttt{\textbf{joints}} vs. \texttt{\textbf{skeleton}} and \texttt{\textbf{mesh}}). Learning-based techniques with \texttt{\textbf{mesh}} features that operate on pointclouds are able to outperform heuristics, even though the latter has access to the full high-resolution object mesh, while the former makes predictions on a pointcloud. Learning also enables \texttt{\textbf{skeleton}} features, which have access to only the 3D joint locations, to perform competitively against mesh-based heuristics and features. While image-based techniques are not yet as accurate as the hand pose-based ones, a significant boost is achieved with multi-view inputs.

\begin{figure}
    \includegraphics[width=\textwidth]{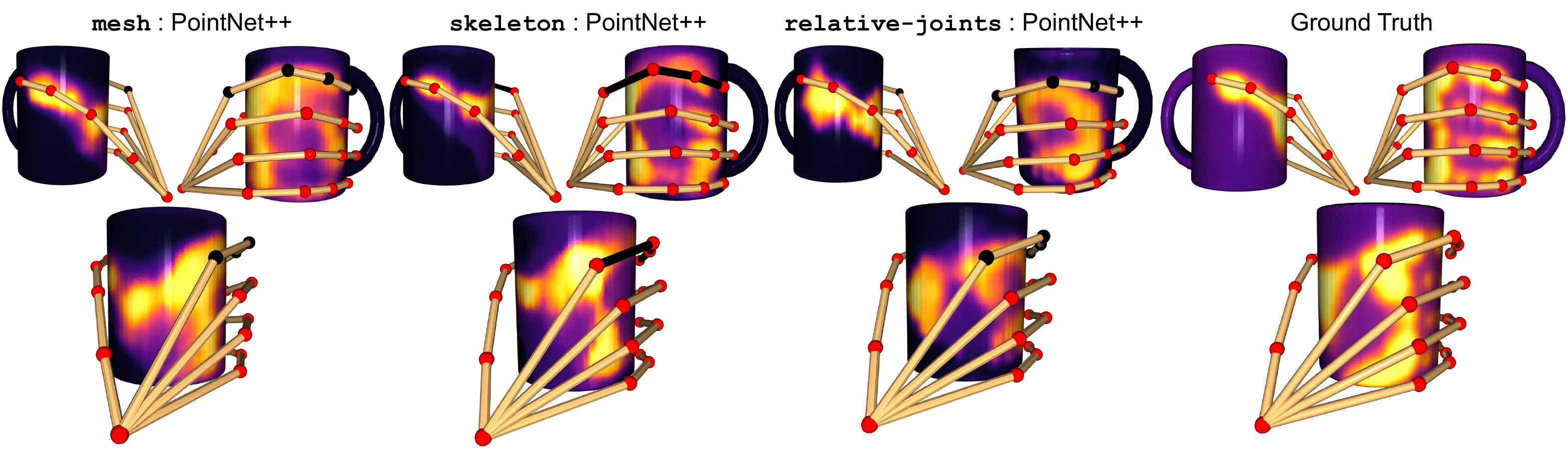}
    \caption{Contact prediction for mug (an unseen object) from hand pose. All input features related to black line segments and joints were dropped (set to 0). Notice how the \textbf{\texttt{mesh}}- and \textbf{\texttt{skeleton}}-PointNet++ predictors is able to capture nuances of palm contact, thumb and finger shapes.}
    \label{fig:pred_from_hand_pose}
  \end{figure}

Figure \ref{fig:pred_from_hand_pose} shows contact prediction results from hand pose for mug, an unseen object. Predictions are transferred from the pointcloud to high-resolution meshes for better visualization. The \textbf{\texttt{skeleton}}-PointNet++ combination is able to predict plausible contact patterns for dropped-out parts of the hand, and capture some of the nuances of palm contact. The \textbf{\texttt{mesh}}-PointNet++ combination captures more nuances, especially at the thumb and bottom of the palm. In contrast, \textbf{\texttt{relative-joints}} features-based predictions are diffused, lack finer details, and have high contact probability in the gaps between fingers, possibly due to lack of access to information about joint connectivity and hand shape.

\begin{figure}
\begin{subfigure}[b]{0.63\textwidth}
    \includegraphics[width=\textwidth]{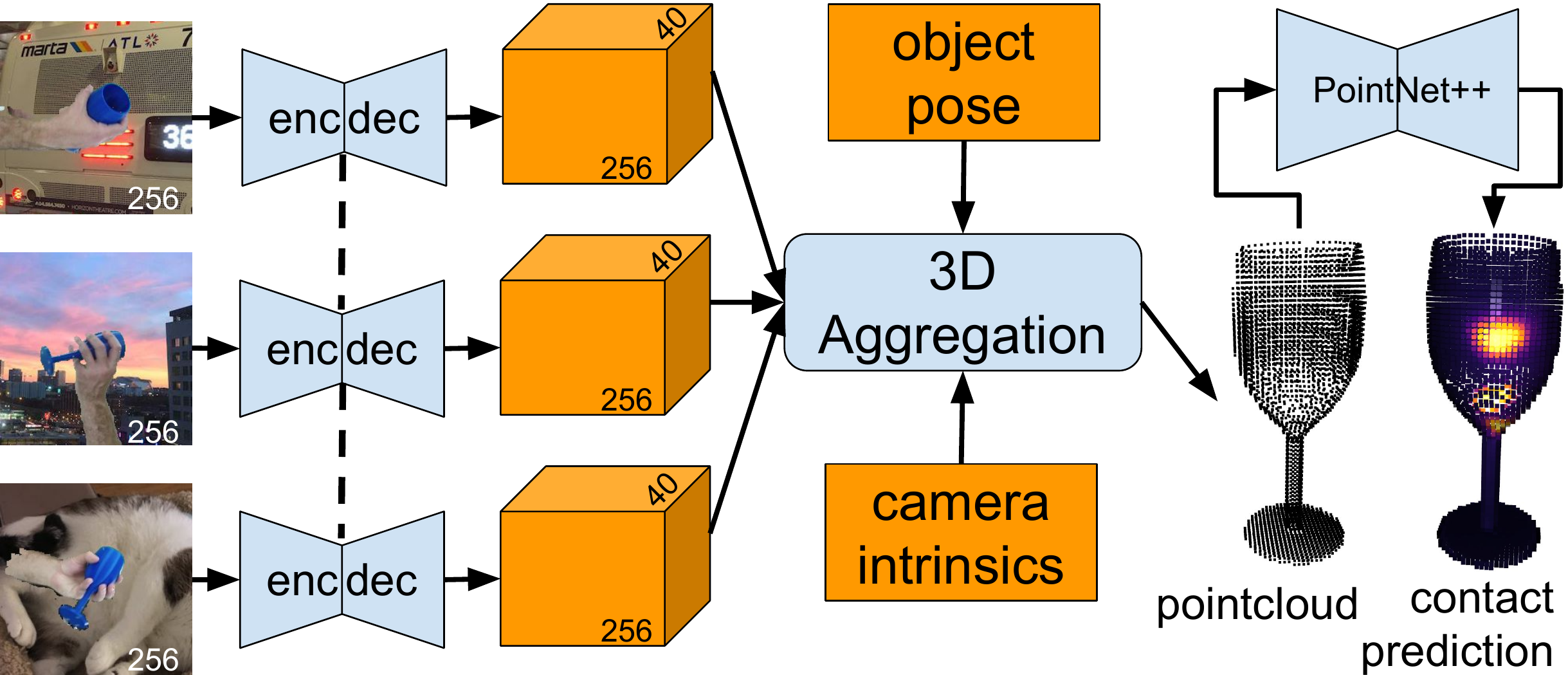}
\caption{}
\label{fig:image_pred_architecture} 
\end{subfigure}
\begin{subfigure}[b]{0.35\textwidth}
    \includegraphics[width=\textwidth]{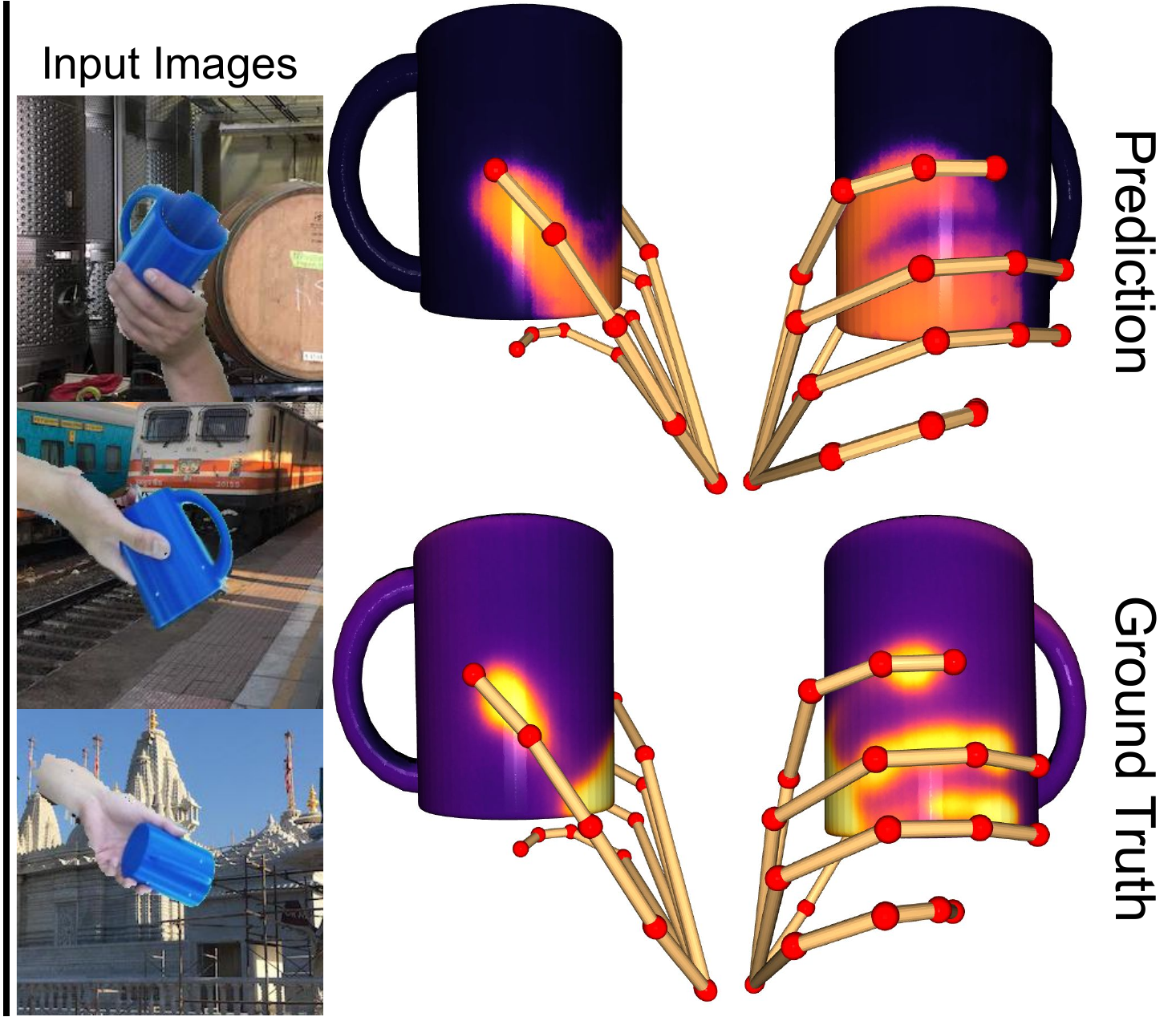}
\caption{}
\label{fig:pred_from_images} 
\end{subfigure}
\caption{(a) Image-based contact prediction architecture. (b) Contact prediction for mug (an unseen object) from RGB images, using networks trained with 3 views. Hand poses shown only for reference.}
\end{figure}

Figure \ref{fig:pred_from_images} shows contact prediction results from RGB images for mug, an unseen object. These predictions have less high-frequency details compared to hand pose based predictions. They also suffer from depth ambiguity -- the proximal part of the index finger appears to be in contact from the mug images, but is actually not. This can potentially be mitigated by use of depth images.
\section{Conclusion and Future Work}
We introduced ContactPose, the first dataset of paired hand-object contact, hand pose, object pose, and RGB-D images for functional grasping. Data analysis revealed some surprising patterns, like higher concentration of hand contact at the first three fingers for `hand-off' vs. `use' grasps. We also showed how learning-based techniques for geometry-based contact modeling can capture nuanced details missed by heuristic methods.

Using this contact ground-truth to develop more realistic, deformable hand mesh models could be an interesting research direction. State-of-the-art models (\eg \cite{romero2017embodied,joo2018total}) are rigid, while the human hand is covered with soft tissue. As the Future Work section of \cite{romero2017embodied} notes, they are trained with meshes from which objects are manually removed, and do not explicitly reason about hand-object contact. ContactPose data can potentially help in the development and evaluation of hand mesh deformation algorithms.

\FloatBarrier
\noindent\textbf{Acknowledgements}: We are thankful to the anonymous reviewers for helping improve this paper. We would also like to thank Elise Campbell, Braden Copple, David Dimond, Vivian Lo, Jeremy Schichtel, Steve Olsen, Lingling Tao, Sue Tunstall, Robert Wang, Ed Wei, and Yuting Ye for discussions and logistics help.
\appendix
\section*{Supplementary Material}

\begin{abstract}
The supplementary material includes a discussion on contact capture, accuracy evaluation of the hand pose and contact ground truth, MANO hand mesh \cite{romero2017embodied} fitting details, network architectures, and implementation details for the learning algorithms. Finally, we present the list of objects and their `use' instructions, and describe the participants' hand information that is included in ContactPose. Please see the extended supplementary material at \url{https://contactpose.cc.gatech.edu} for example RGB-D imagery and slices through the data in the form of 1) object- and intent-specific hand contact probabilities, and 2) `use' vs. `hand-off' contact maps and hand poses for all grasps of an object.
\end{abstract}

\section{Contact Capture Discussion}
The process to convert thermal image pixels to contact values follows \cite{contactdbv1}. Raw thermal readings were converted to continuous contact values in $[0, 1]$ using a sigmoid that maps the warmest point to $0.95$ and the coldest point to $0.05$. These values non-linearly encode the temperature of the object, where $[0, 1]$ approximately corresponds to [room temperature, body temperature]. While most experiments used this continuous value, if a hard decision about the contact status of a point was desired, this was done by thresholding these processed values at $0.4$.
\section{MANO Fitting}
This section provides details for the fitting procedure of the MANO \cite{romero2017embodied} hand model to ContactPose data. Borrowing notation from \cite{romero2017embodied}, the MANO model is a mesh with vertices $M\left(\beta, \theta\right)$ parameterized by shape parameters $\beta$ and pose parameters $\theta$. The 3D joint locations of the posed mesh, denoted here by $J\left(\beta, \theta\right)$, are also a function of the shape and pose parameters. We modify the original model by adding one joint at each fingertip, thus matching the format of joints $J^*$ in ContactPose annotations.

MANO fitting is performed by optimizing the following objective function, which combines L2 distance of 3D joints and shape parameter regularization:

\begin{equation}
    \beta^*, \theta^* = \arg min_{\beta, \theta} \vert\vert J\left(\beta, \theta\right) - J^* \vert\vert + \frac{1}{\sigma} \vert\vert\beta\vert\vert
\end{equation}

where $\sigma$ is set to 10. It is optimized using the Dogleg \cite{powell1970new} optimizer implemented in chumpy \cite{chumpy}. We initialized $\beta$ and $\theta$ to $\mathbf{0}$ (mean shape and pose) after 6-DOF alignment of the wrist and 5 palm joints. Finally, the MANO model includes a PCA decomposition of 45 pose parameters to 6 parameters by default. We provide MANO fitting data with 10 and 15 pose components in the ContactPose dataset, but use the MANO models with 10 pose components in all our contact modeling experiments.

\section{Dataset Accuracy}
In this section, we cross-evaluate the accuracy of the hand pose and contact data included in ContactPose.

\subsection{Contact Accuracy}
We note that conduction is the principal mode of heat transfer in solid-to-solid contact \cite{ahtt5e}. Combined with the observation by Brahmbhatt \etal \cite{contactdbv1} that heat dissipation within the 3D printed objects is low over the time scales we employ to scan them, this indicates that conducted heat can accurately encode contact. Following \cite{contactdbv1}, we measure the conducted heat with a thermal camera.

\noindent\textbf{Agreement with MANO Hand Mesh}: The average distance of contacted object points from their nearest hand point is 4.17 mm (10 MANO hand pose parameters) and 4.06 mm (15 MANO hand pose parameters).

\begin{figure}
  \includegraphics[width=\columnwidth]{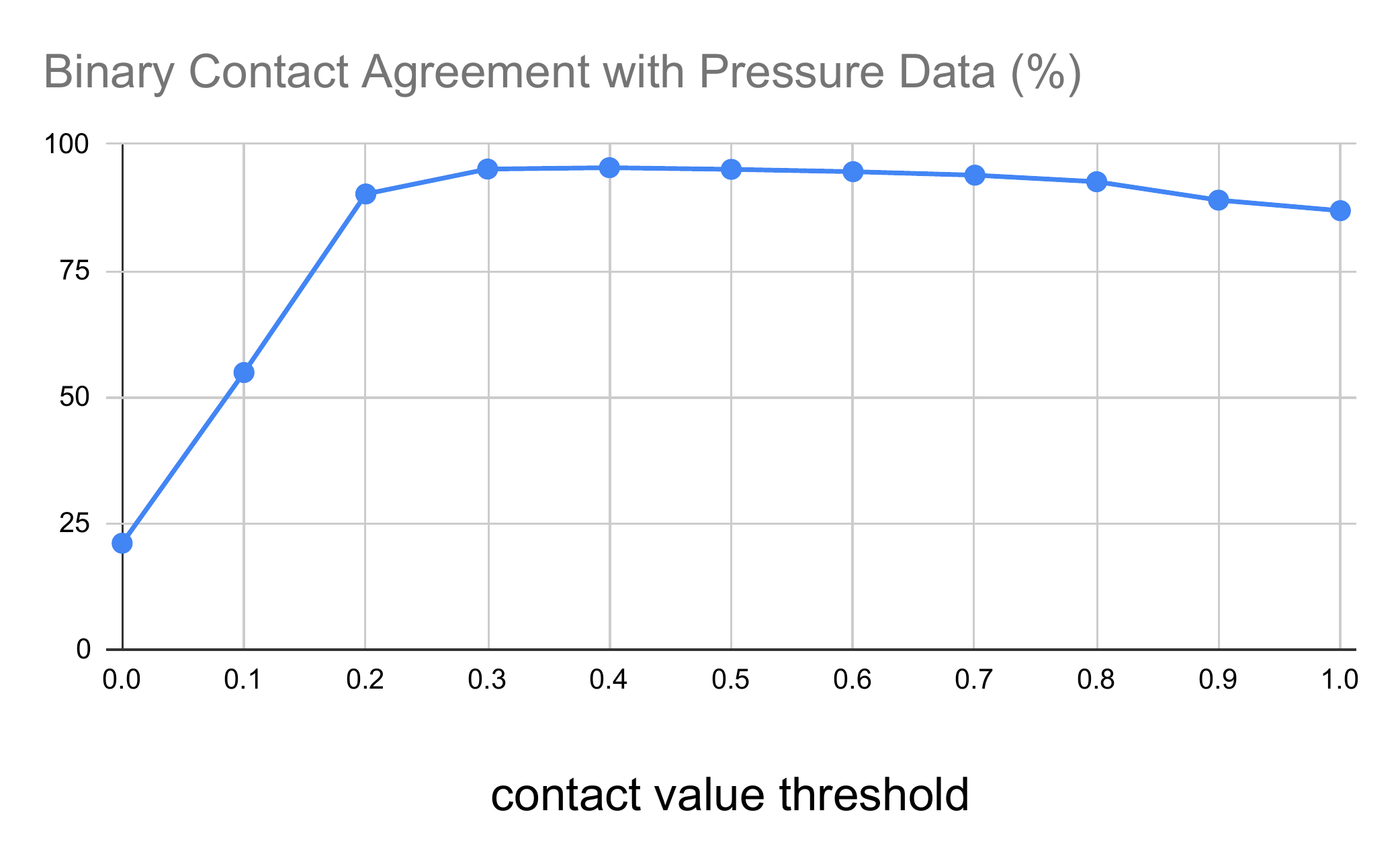}
  \caption{Relation of contact value threshold to the binary contact agreement with pressure data from the Sensel Morph sensor. Agreement maximizes at the
  threshold value of 0.4.}
  \label{fig:contact_verification}
\end{figure}

\noindent\textbf{Agreement with Pressure-based Contact}: We also verified thermal contact maps against pressure images from a Sensel Morph planar pressure sensor \cite{sensel_morph_website, sensel_morph_patent}. After registering the thermal and pressure images, we thresholded the processed thermal image at values in $[0, 1]$ with an interval of 0.1. Any nonzero pixel in the pressure image is considered to be contacted. Binary contact agreement peaks at 95.4\% at the threshold of 0.4 (Figure \ref{fig:contact_verification}).

\subsection{3D Hand Pose Accuracy}
Following \cite{hampali2019ho}, this is measured through the discrepancy between 3D joints of the fitted MANO model and the ground truth 3D joints. Low-quality physically implausible ground truth can yield higher discrepancy, since the MANO model is not able to fit to physically implausible joint locations. Table \ref{tab:joint_accuracy} shows that ContactPose has significantly lower discrepancy than other recent datasets, even though it uses less than one-third MANO hand pose parameters. Table \ref{tab:penetration_stats} shows statistics for hand-object penetration.
\begin{table}
    \centering
    \resizebox*{\textwidth}{!}{
    \begin{tabular}{c|c|c}
        \textbf{Dataset} & \textbf{Avg. 3D Joint Error (mm)} & \textbf{AUC (\%)}\\\hline
        HO-3D \cite{hampali2019ho} & 7.7 & 79.0\\
        FreiHand \cite{Freihand2019} & - & 79.1\\
        HANDS 2019 \cite{hands2019} & 11.39 & -\\
        ContactPose (ours) -- 10 pose params & 7.65 & 84.54\\
        ContactPose (ours) -- 15 pose params & \textbf{6.68} & \textbf{86.49}\\
    \end{tabular}}
    \caption{Discrepancy between 3D joints of the fitted MANO model and the ground truth 3D joints. 3D joint error (lower is better) is averaged over all 21 joints. AUC (higher is better) is the area under the error threshold vs. percentage of correct keypoints (PCK) curve, where the maximum error threshold is set to 5 cm.}
    \label{tab:joint_accuracy}
\end{table}

\begin{table}
    \centering
    \resizebox*{\textwidth}{!}{
    \begin{tabular}{c|c|c|c}
        \textbf{Dataset} & \textbf{Mean Penetration (mm)} & \textbf{Median Penetration (mm)} & \textbf{Penetration freq (\%)}\\\hline
        FPHA \cite{FHAD_FirstPersonAction} (reported in \cite{learningJointReconstructionOfHandsAndManipulatedObjects}) & 11.0 & - & -\\
        ContactPose -- 15 pose params & \textbf{2.02} & 1.53 & 4.75\\
    \end{tabular}}
    \caption{Statistics for hand-object penetration showing the accuracy of ContactPose. Note that \cite{learningJointReconstructionOfHandsAndManipulatedObjects} report \emph{joint} penetration for \cite{FHAD_FirstPersonAction}, while we report \emph{surface} penetration.}
    \label{tab:penetration_stats}
\end{table}
\section{Participants' Hand Information}
\begin{figure}[h!]
  \includegraphics[width=0.35\columnwidth]{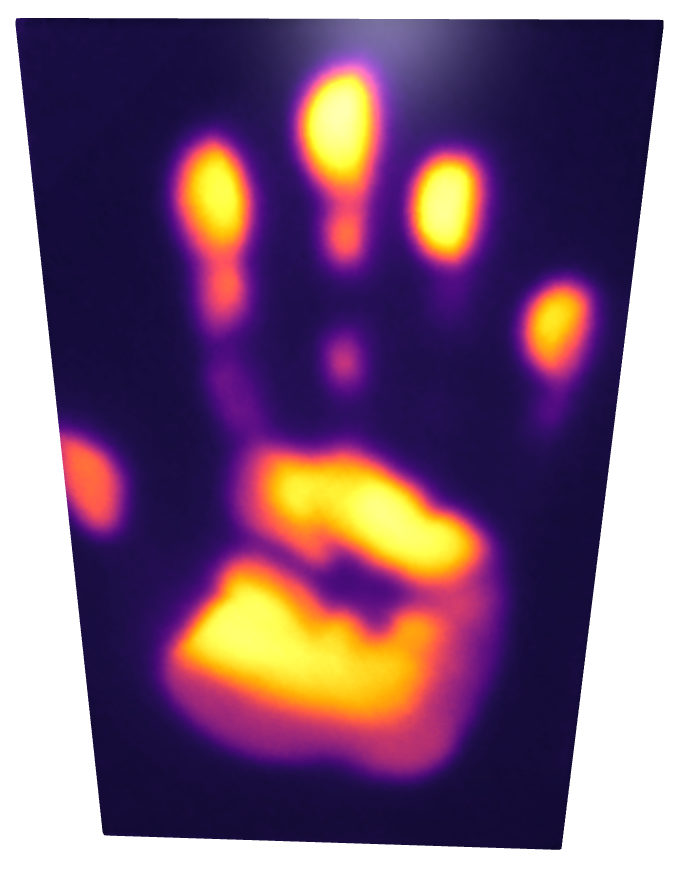}
  \caption{Contact map of a participant's palm on a flat plate. Such palm contact maps for each participant are included in ContactPose.}
  \label{fig:hand_profile_plate}
\end{figure}

\begin{figure}[h!]
  \includegraphics[width=0.5\columnwidth]{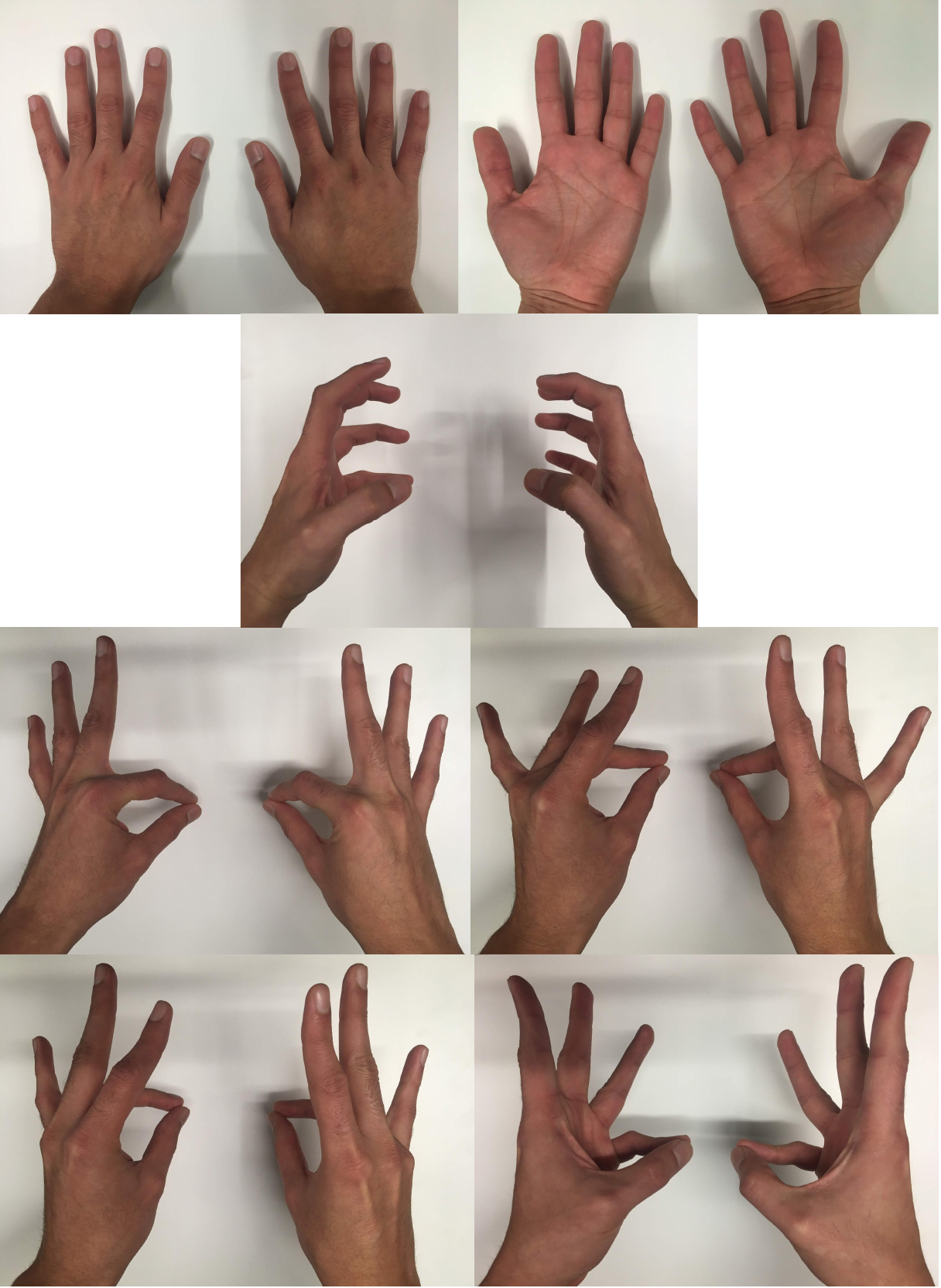}
  \caption{Pre-defined hand gestures performed by each participant. RGB-D videos from 3 Kinects of each participant performing these gestures are included in ContactPose.}
  \label{fig:hand_profile_gestures}
\end{figure}

We captured information about each ContactPose participant's hands in two ways: 1) contact map on a flat plate (example shown in Figure~\ref{fig:hand_profile_plate}), and 2) RGB-D videos of the participants performing 7 hand gestures (shown in Figure~\ref{fig:hand_profile_gestures}). This can potentially be used to estimate the hand shape by fit embodied hand models (e.g.~\cite{romero2017embodied}).

\section{Network Architectures}
\subsection{PointNet++}
The PointNet++ architecture we use is similar to the pointcloud segmentation network from Qi et al~\cite{qi2017pointnet++}, with modifications aiming to reduce the number of learnable parameters. Similarly to~\cite{qi2017pointnet++}, we use $SA\left(s, r, [l_1, \ldots, l_d]\right)$ to indicate a Set Abstraction layer with a farthest point sampling ratio $s$, ball radius $r$ (the pointcloud is normalized to lie in the $[-0.5, 0.5]$ cube) and $d$ fully connected layers of size $l_i (i=1 \ldots d)$. The global Set Abstraction layer is denoted without farthest point sampling ratio and ball radius. $FP\left(K, [l_1, \ldots, l_d]\right)$ indicates a Feature Propagation layer with $K$ nearest neighbors and $d$ fully connected layers of size $l_i (i=1 \ldots d)$. $FC\left(S_{in}, S_{out}\right)$ indicates a fully connected layer of output size $S_{out}$ applied separately to each point (which has $S_{in}$-dimensional features). Each fully connected layer in the Set Abstraction and Feature Propagation layers is followed by ReLU and batch-norm layers. Our network architecture is:
\begin{align*}
    &SA\left(0.2, 0.1, [F, 64, 128]\right) - SA\left(0.25, 0.2, [128, 128, 256]\right) - \\
    &SA\left([256, 512, 1024]\right) - FP(1, [1024 + 256, 256, 256]) - \\
    &FP(3, [256 + 128, 256, 128]) - FP(3, [128+F, 128, 128]) - \\
    &FC(128, 128) - FC(128, 10)
\end{align*}
where $F$ is the number of input features and the final layer outputs scores for the 10 contact value classes.

\subsection{Image Encoder-Decoder}
\begin{figure}
  \includegraphics[width=\columnwidth]{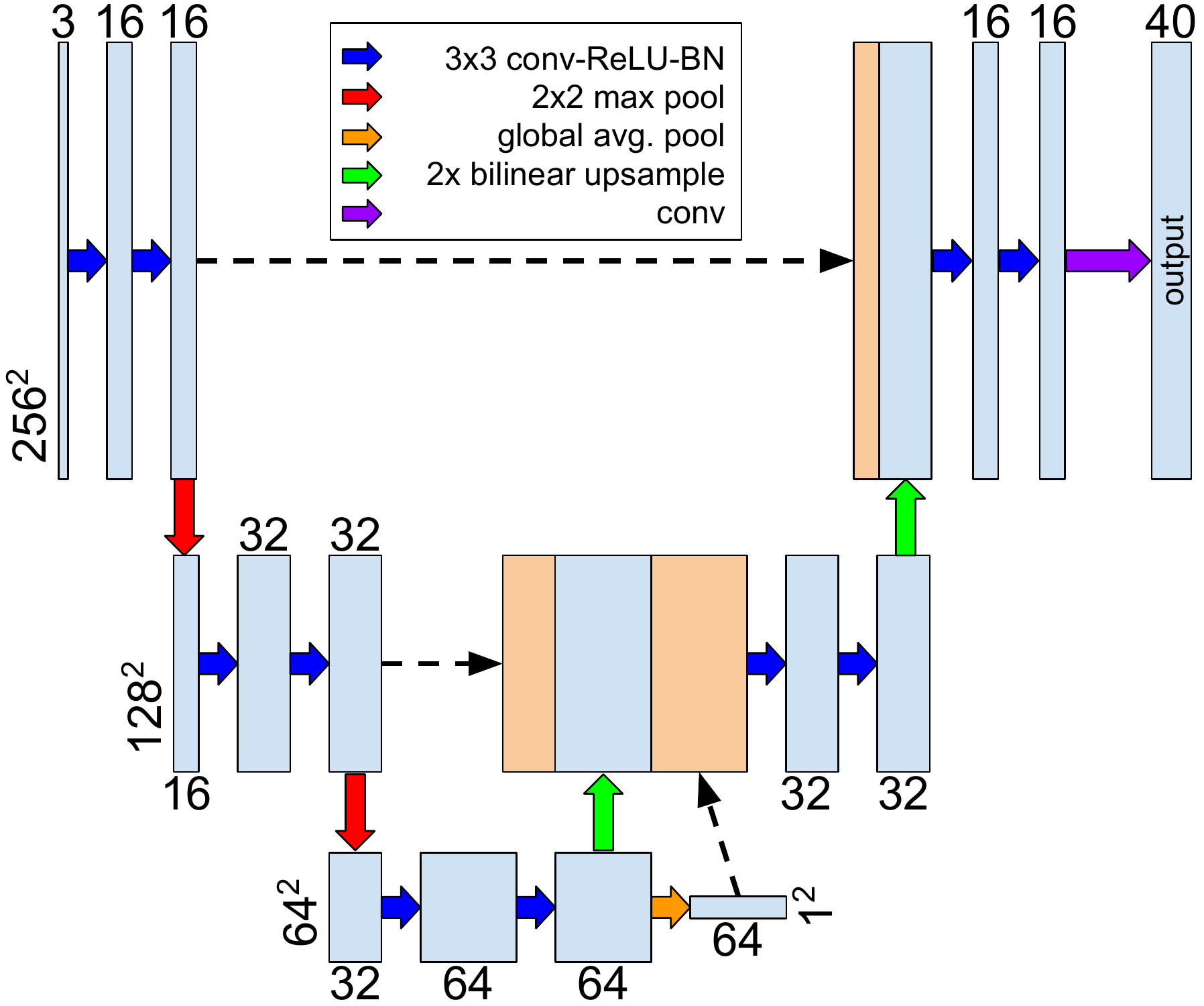}
  \caption{Architecture for the image encoder-decoder (Figure 10 in main paper). Horizontal numbers indicate number of channels, and vertical numbers indicate spatial dimensions.}
  \label{fig:enc_dec_architecture}
\end{figure}
We take inspiration from U-Net~\cite{ronneberger2015unet} and design the light-weight network shown in~\ref{fig:enc_dec_architecture} that extracts dense features from RGB images. The global average pooling layer is intended to capture information about the entire hand and object.

\section{Training and Evaluation Details}
All models are trained using PyTorch~\cite{paszke2017automatic} and the Adam optimizer~\cite{kingma2014adam} (base learning rate $\in \{5\times10^{-4}, 1\times10^{-3}, 5\times10^{-3}\}$, momentum of 0.9, weight decay of $5e-4$, and a batch size of 25). Both point-clouds and voxel-grids are rotated around their `up'-axis at regularly spaced $30\degree$ intervals. These rotations are considered separate data points during training, and their predictions are averaged during evaluation.

For image-based contact prediction, ContactPose has approximately 300 RGB-D frames ($\times$ 3 Kinects) for each grasp, but temporally nearby frames are highly correlated because of the high frame rate. Hence, we include equally spaced 50 frames from each grasp in the training set. Evaluation is performed over equally spaced 12 frames from this set of 50 frames.
\section{List of Objects}
Table~\ref{tab:object_list} shows a list of all 25 objects in ContactPose, along with information about the which of these objects are included in the two functional grasping 
categories, and the specific `use' instructions.

\begin{table*}
\centering
\begin{tabular}{c|c|c|c}
\textbf{Object} & \textbf{handoff} & \textbf{use} & \textbf{use instruction}\\
\hline
apple & \checkmark & \checkmark & eat\\
banana	& \checkmark & \checkmark & peel\\
binoculars & \checkmark & \checkmark & see through\\
bowl & \checkmark & \checkmark & drink from\\
camera & \checkmark & \checkmark & take picture\\
cell phone & \checkmark & \checkmark & talk on\\
cup & \checkmark & \checkmark & drink from\\
door knob	& & \checkmark & twist to open door\\
eyeglasses & \checkmark & \checkmark & wear\\
flashlight & \checkmark & \checkmark & turn on\\
hammer & \checkmark & \checkmark & hit a nail\\
headphones & \checkmark & \checkmark & wear\\
knife & \checkmark & \checkmark & cut\\
light bulb & \checkmark & \checkmark & screw in a socket\\
mouse & \checkmark	 & \checkmark & use to point and click\\
mug & \checkmark & \checkmark & drink from\\
pan & \checkmark & \checkmark & cook in\\
PS controller & \checkmark & \checkmark & play a game with\\
scissors & \checkmark & \checkmark & cut with\\
stapler & \checkmark & \checkmark & staple\\
toothbrush & \checkmark & \checkmark & brush teeth\\
toothpaste & \checkmark &\checkmark & squeeze out toothpaste\\
Utah teapot & \checkmark & \checkmark & pour tea from\\
water bottle & \checkmark & \checkmark & open\\
wine glass	 & \checkmark & \checkmark & drink wine from\\
\hline
\textbf{Total} & \textbf{24} & \textbf{25} &\\
\end{tabular}
\caption{List of objects in ContactPose and specific `use' instructions}
\label{tab:object_list}
\end{table*}

\balance
\bibliographystyle{splncs04}
\bibliography{references}
\end{document}